\def\year{2021}\relax
\documentclass[letterpaper]{article} % DO NOT CHANGE THIS
\usepackage{aaai21}  % DO NOT CHANGE THIS
\usepackage{times}  % DO NOT CHANGE THIS
\usepackage{helvet} % DO NOT CHANGE THIS
\usepackage{courier}  % DO NOT CHANGE THIS
\usepackage[hyphens]{url}  % DO NOT CHANGE THIS
\usepackage{graphicx} % DO NOT CHANGE THIS
\usepackage{natbib}  % DO NOT CHANGE THIS AND DO NOT ADD ANY OPTIONS TO IT
\usepackage{caption} % DO NOT CHANGE THIS AND DO NOT ADD ANY OPTIONS TO IT
\usepackage{amsmath}
\usepackage{bm}
\usepackage{mathrsfs}
\usepackage{amsthm}
\usepackage{amsfonts}
\usepackage[ruled,vlined]{algorithm2e}
\usepackage{algorithmic}
\usepackage{bm}
\usepackage{subfigure}
\usepackage{multirow}
\usepackage{url}
\usepackage{array}

\newcommand{\ie}{{\sl i.e.}}
\newcommand{\eg}{{\sl e.g.}}

\newcommand{\wrt}{{\sl w.r.t. }}
\newcommand{\aka}{{\sl a.k.a. }}

\title{Hyperbolic Variational Graph Neural Network for Modeling Dynamic Graphs}
\author{
    Li Sun\textsuperscript{\rm 1}, 
    Zhongbao Zhang\textsuperscript{\rm 1}\thanks{Corresponding Authors}, 
    Jiawei Zhang\textsuperscript{\rm 2}, 
    Feiyang Wang\textsuperscript{\rm 1}, 
    Hao Peng\textsuperscript{\rm 3}, \\
    Sen Su\textsuperscript{\rm 1*} and 
    Philip S. Yu\textsuperscript{\rm 4}\\
}

\affiliations{
    \textsuperscript{\rm 1}State Key Laboratory of Networking and Switching Technology, \\Beijing University of Posts and Telecommunications, China\\
    \textsuperscript{\rm 2}IFM Lab, Department of Computer Science, Florida State University, FL, USA\\
    \textsuperscript{\rm 3}Beijing Advanced Innovation Center for Big Data and Brain Computing, Beihang University, China\\
    \textsuperscript{\rm 4}Department of Computer Science, University of Illinois at Chicago, IL, USA\\
    \{l.sun, zhongbaozb, fywang\}@bupt.edu.cn; jiawei@ifmlab.org; penghao@act.buaa.edu.cn; psyu@uic.edu

}

\begin{document}
%\linenumbers
\maketitle

\begin{abstract}
Learning representations for graphs plays a critical role in a wide spectrum of downstream applications.
In this paper, we summarize the limitations of the prior works in three folds: representation space, modeling dynamics and modeling uncertainty.
To bridge this gap, we propose to learn dynamic graph representation in hyperbolic space, for the first time, which aims to infer stochastic node representations.
Working with hyperbolic space, we present a novel Hyperbolic Variational Graph Neural Network, referred to as HVGNN.
In particular, 
to model the dynamics,
we introduce a Temporal GNN (TGNN) based on a theoretically grounded time encoding approach.
To model the uncertainty,
we devise a hyperbolic graph variational autoencoder built upon the proposed TGNN to generate stochastic node representations of hyperbolic normal distributions.
Furthermore, we introduce a reparameterisable sampling algorithm for the hyperbolic normal distribution to enable the gradient-based learning of HVGNN.
Extensive experiments show that HVGNN outperforms state-of-the-art baselines on real-world datasets.
\end{abstract}

%!TEX root = ./main.tex

\section{Introduction}

Recent years have witnessed a surge of representation learning on graphs \cite{DeepWalk,tang2015line,Xhonneux2020CGNN}.
Its basic idea is to map each node to a vector in a low-dimensional representation space.
By learning graph representations, classical machine learning algorithms can be applied to solve various graph analysis tasks, such as link prediction and node classification \cite{kipf2016semi}.

%Modeling dynamic graphs is challenging, and i
% It is not until recently that several solutions \cite{kumar2019predicting,Xhonneux2020CGNN} are proposed. 
In this paper, we summarize the limitations of the prior graph representation learning works in three folds:

\noindent \textbf{1) \emph{Representation Space.}}
Most of existing studies \cite{kipf2016variational,Xu2020inductive} model graphs in the Euclidean space. 
Euclidean models tend to have distortions when representing real-world graphs with latent hierarchies \cite{HGCN,chen2013hyperbolicity}.
In particular, for such graphs, the number of nodes surrounding to a center node grows \emph{exponentially} \wrt radius. 
However, the size of the Euclidean space only grows polynomially \wrt radius,
while this size grows \emph{exponentially} in hyperbolic space \cite{krioukov2010hyperbolic}.
\emph{Hyperbolic} space provides a more promising alternative.
Actually, recent results \cite{papadopoulos2012popularity,HGNN} show that hyperbolic space is well-suited for modeling graphs.

\noindent \textbf{2) \emph{Modeling Dynamics.}}
% To date, representation learning on graphs has been mostly studied in static
% settings while efforts for modeling dynamic graphs are still limited.
Most of existing models \cite{2020Learning,tu2018deep,wang2016structural} consider the graphs to be static.
%However, real-world problems are naturally modeled by \emph{dynamic} graphs, 
Actually, graphs are usually \emph{dynamic} and constantly evolving over time.
%where graphs are 
Dynamic graphs have been typically observed in social networks, transportation networks and financial transaction networks \cite{zhou2018dynamic}.
Ignoring the inherent dynamics of graphs usually leads to questionable inference. 
Such models may mistakenly utilize future information for predicting past interactions as the evolving constraints are disregarded.
% Though a few hyperbolic models for static graphs \cite{HGNN,HAN} have been proposed,
% hyperbolic models for dynamic graphs have not been touched yet.

	%real-world graphs are much more complex than we assume.  

\noindent \textbf{3) \emph{Modeling Uncertainty.}}
Most of existing models \cite{liu2020fine,node2vec} map nodes to deterministic vectors.
		% Meanwhile, recent studies \cite{HGNN,HAN} report more favorable performance on exploring hyperbolic space to model graph structure than Euclidean space.
	% However, they have not yet incorporate the graph dynamics.
	% We are to propose a hyperbolic model for dynamic graphs at the first attempt.
However, the formation and evolution of graphs are full of uncertainties, especially for low-degree nodes which deliver less information and bear more uncertainties \cite{zhu2018dvne}.
Actually, uncertainty is an inherent characteristic of graphs.
The deterministic representation cannot model uncertainty.
Alternatively, \emph{stochastic} representation provides a promising approach to model such characteristic. 
It naturally captures the uncertainty to represent nodes as normal distributions, \ie, the mean and variance.
%, rather than deterministic vectors. 

To address the aforementioned limitations, 
we propose to learn dynamic graph representation in hyperbolic space, for the first time, 
which aims to learn stochastic node representations modeling graph dynamics and its uncertainty.

To this end,  we present a novel Hyperbolic Variational Graph Neural Network, referred to as HVGNN.
In HVGNN, 
to address the first limitation,
instead of the Euclidean space, we utilize hyperbolic space as the representation space.
To address the second limitation, 
we introduce a novel Temporal GNN (TGNN) to model the dynamics. 
In particular, TGNN performs a time-aware attention based on a theoretically grounded time encoding approach, which distinguishes nodes in time domain.
To address the third limitation,
we devise a hyperbolic graph variational autoencoder built upon TGNN to jointly model the uncertainty and dynamics.
In particular, we generate stochastic representations of wrapped normal distributions, the generalized normal distributions in hyperbolic space, whose parameters are defined by TGNN.
Furthermore, we introduce a reparameterisable sampling algorithm for wrapped normal distributions to enable the gradient-based learning of HVGNN.

We evaluate HVGNN by two typical downstream tasks on graph data: link prediction and node classification.
Extensive experiments on real-world datasets show that HVGNN outperforms several state-of-the-art methods.

Overall, main contributions of our work are listed below:
\begin{itemize}
	\item To the best of our knowledge, this is the first attempt to learn the node representations for dynamic graphs in hyperbolic space.
	\item We propose a novel Hyperbolic Variational Graph Neural Network, HVGNN, which generates stochastic representations to model graph dynamics and its uncertainty in hyperbolic space.
	%\item Extensive experiments demonstrate the superiority of HVGNN on several real-world datasets.
	\item Experimental results show the superiority of HVGNN on several real-world datasets.
	%We demonstrate the superiority of the proposed approach through extensive experiments on real-world datasets.
\end{itemize}

%!TEX root = ./main.tex

\section{Preliminaries: Hyperbolic Geometry}
%Prior to introducing the studied problem and proposed approach, we provide some important preliminaries about hyperbolic geometry in this section. 
For the in-depth expositions, refer mathematical materials \cite{Spivak1979,Hopper2010}.
Throughout the paper, we denote the Euclidean norm and inner product by $\| \cdot \|$ and $\langle \cdot, \cdot \rangle$, respectively.

\subsection{Riemannian Manifold}
% For a rigorous reasoning about hyperbolic spaces, we briefly introduce the concepts in Riemannian geometry largely needed in this paper. 

% Mapping between the tangent space and the manifold is done via exponential and logarithmic maps.
% For a point  $\mathbf x \in \mathcal M$ in the manifold, the exponential map at $\mathbf x$, $\exp_\mathbf x(\mathbf v)$, defines the map from  tangent space $\mathcal T_\mathbf x\mathcal M$ to the manifold $\mathcal M$ that projects the vector $\mathbf v \in \mathcal T_\mathbf x\mathcal M$ onto $\mathcal M$.
% The logarithmic map at $\mathbf x$, $\log_\mathbf x(\mathbf y)$, projects the vector $\mathbf y \in \mathcal M$ back to the tangent space $\mathcal T_\mathbf x\mathcal M$.
% Mapping between tangent spaces is done via parallel transport.
% For two points  $\mathbf x, \mathbf y \in \mathcal M$ in the manifold, the parallel transport from $\mathbf x$ to $\mathbf y $, $P_{\mathbf x \to \mathbf y}(\mathbf v)$, defines the map from tangent space $\mathcal T_\mathbf x\mathcal M$ to  $\mathcal T_\mathbf y\mathcal M$ that carries the vector $\mathbf v \in \mathcal T_\mathbf x\mathcal M$ to $\mathcal T_\mathbf y\mathcal M$ along the geodesic.

A \emph{manifold} $\mathcal M$ is a space that generalizes the notion of a 2D surface to higher dimensions \cite{HNN}.
For each point $\mathbf x \in \mathcal M$, it associates with a \emph{tangent space} $\mathcal T_\mathbf x\mathcal M$ of the same dimensionality as $\mathcal M$.
Intuitively, $\mathcal T_\mathbf x\mathcal M$ contains all possible directions in which one can pass through $\mathbf x$ tangentially.
On the associated tangent space $\mathcal T_\mathbf x\mathcal M$, the \emph{Riemannian metric}, $g_\mathbf x (\cdot, \cdot) : \mathcal T_\mathbf x\mathcal M  \times \mathcal T_\mathbf x\mathcal M \to \mathbb R$, defines an inner product specifying the geometry of $\mathcal M$.
% locally, from which the global quantities can be inferred.
% A manifold $\mathcal M$ equipped with a Riemannian metric $g$ is defined as a \emph{Riemannian manifold} $(\mathcal M, g)$.
A \emph{Riemannian manifold} is then defined as the tuple of $(\mathcal M, g)$.

% A smooth path on the manifold of minimal length between two points is called the geodesic, and can be seen as the generalization of a straight-line in Euclidean space.

Mapping between the tangent space and the manifold is done via exponential and logarithmic maps.
For a point  $\mathbf x \in \mathcal M$ in the manifold, the  \emph{exponential map} at $\mathbf x$, $\exp_\mathbf x(\mathbf v): \mathcal T_\mathbf x\mathcal M \to \mathcal M$, projects the vector $\mathbf v \in \mathcal T_\mathbf x\mathcal M$ onto the manifold $\mathcal M$.
The \emph{logarithmic map} at $\mathbf x$, $\log_\mathbf x(\mathbf y): \mathcal M \to \mathcal T_\mathbf x\mathcal M$, projects the vector $\mathbf y \in \mathcal M$ back to the tangent space $\mathcal T_\mathbf x\mathcal M$.
Mapping between tangent spaces is done via parallel transport.
For two points  $\mathbf x, \mathbf y \in \mathcal M$ in the manifold, the \emph{parallel transport} from $\mathbf x$ to $\mathbf y $, $P_{\mathbf x \to \mathbf y}(\mathbf v): \mathcal T_\mathbf x\mathcal M \to \mathcal T_\mathbf y\mathcal M$, carries the vector $\mathbf v \in \mathcal T_\mathbf x\mathcal M$ to $\mathcal T_\mathbf y\mathcal M$ along the \emph{geodesic}, a smooth path on the manifold of minimal length between  $\mathbf x$ and $\mathbf y$.

As one of the fundamental objects, the hyperbolic space is a Riemannian manifold with constant negative curvature.
There are four common equivalent models of hyperbolic space: the Poincar\'{e} disk model, Poincar\'{e} half-plane model, Lorentz (\aka hyperboloid/Minkowski) model and Klein model \cite{HGNN}.
Next, we give more details of the latter two, which will both be utilized in this paper, for further discussions.

\subsection{The Lorentz Model}
We denote $\mathbb L^{d, K}$ as the Lorentz model in $d$ dimensions with constant negative  \emph{curvature} $-1/K$ where $K > 0$. 
The Lorentz model $\mathbb L^{d, K}$ is defined on a subset of $\mathbb R^{d+1}$, $\mathbb L^{d, K}=\{\mathbf x \in \mathbb R^{d+1} | \langle \mathbf x, \mathbf x \rangle_{\mathcal L} = -K\}$, 
where $\langle \cdot, \cdot \rangle_{\mathcal L}: \mathbb{R}^{d+1} \times \mathbb{R}^{d+1}\rightarrow\mathbb{R}$ is Lorentzian inner product defined as,
\vspace{-0.03in}
\begin{equation}
\langle \mathbf x, \mathbf y \rangle_{\mathcal L}= -x_0y_0+x_1y_1+x_2y_2+ \dots +x_dy_d.
\label{lproduct}
\end{equation}
We denote $\mathcal T_{\mathbf x}\mathbb L^{d, K}$ as the \emph{tangent space} of $\mathbf x \in \mathbb L^{d, K}$. We have $\mathcal T_{\mathbf x}\mathbb L^{d, K}=\{\mathbf v \in \mathbb R^{d+1} | \langle \mathbf v, \mathbf x \rangle_{\mathcal L} = 0\}$ and $\|\mathbf{v}\|_{\mathcal{L}}=\sqrt{\langle\mathbf{v}, \mathbf{v}\rangle_{\mathcal{L}}}$ is the norm of $\mathbf v$.
For $\mathbf u, \mathbf v \in \mathcal T_{\mathbf x}\mathbb L^{d, K}$, we can give the Riemannian metric tensor $g^K_\mathbf x (\mathbf u, \mathbf v) = \langle \mathbf u, \mathbf v \rangle_{\mathcal L}.$

Next, we give the closed form equations of exponential map, logarithmic map and parallel transport.
For any $\mathbf x, \mathbf y \in \mathbb L^{d, K}$ and $\mathbf v  \in \mathcal T_{\mathbf x}\mathbb L^{d, K}$, we  have the following equations:
\resizebox{1.01\linewidth}{!}{
\begin{minipage}{1.0899\linewidth}
\vspace{-0.07in}
\begin{align}
\exp^K_\mathbf x(\mathbf v)  &= \cosh \left(\frac{\|\mathbf{v}\|_{\mathcal{L}}}{\sqrt{K}}\right) \mathbf{x}+\sqrt{K} \sinh \left(\frac{\|\mathbf{v}\|_{\mathcal{L}}}{\sqrt{K}}\right) \frac{\mathbf{v}}{\|\mathbf{v}\|_{\mathcal{L}}}, \label{exp}
\\
\log^K_\mathbf x(\mathbf y) &=  \frac{d_{\mathcal{L}}^{K}(\mathbf{x}, \mathbf{y})}{\left\|\mathbf{y}+\frac{1}{K}\langle\mathbf{x}, \mathbf{y}\rangle_{\mathcal{L}} \mathbf{x}\right\|_{\mathcal{L}}}   \left(\mathbf{y}+\frac{1}{K}\langle\mathbf{x}, \mathbf{y}\rangle_{\mathcal{L}} \mathbf{x}\right),  \label{log}
\\
P^K_{\mathbf x \to \mathbf y}(\mathbf v) &= \mathbf{v}-\frac{\left\langle\log^K _{\mathbf{x}}(\mathbf{y}), \mathbf{v}\right\rangle_{\mathcal{L}}}{d_{\mathcal{L}}^{K}(\mathbf{x}, \mathbf{y})^{2}}\left(\log^K _{\mathbf{x}}(\mathbf{y})+\log^K _{\mathbf{y}}(\mathbf{x})\right), \label{pt}
\end{align}
\end{minipage}
}
where  $d_{\mathcal{L}}^{K}(\mathbf{x}, \mathbf{y}) = \sqrt{K} \operatorname{cosh}^{-1}\left(-\langle\mathbf{x}, \mathbf{y}\rangle_{\mathcal{L}} / K\right)$. 

\subsection{The Klein Model}
We denote $\mathbb K^{d, K}$ as the Klein model in $d$ dimensions with constant negative curvature $-1/K$ where $K > 0$. 
The Klein model $\mathbb K^{d, K}$ is defined on a subset of $\mathbb R^d$, $\mathbb K^{d, K}=\{\mathbf x \in \mathbb R^d \mid || \mathbf x||^2 < K\}$. 
A point $\mathbf x \in \mathbb K^{d, K}$ is projected from the corresponding $\mathbf y \in \mathbb L^{d, K}$.
We derive the projection with the $i^{th}$  entry:
\vspace{-0.05in}
\begin{equation}
\pi^K_{\mathbb L \to \mathbb K}( y_i) = \sqrt K\frac{ y_i}{ y_0}, 
\vspace{-0.07in}
\label{k2l}
\end{equation}
whose inverse is given as
\vspace{-0.1in}
\begin{equation}
\pi^K_{\mathbb K \to \mathbb L}(\mathbf  x) =\eta^K(\mathbf x)(\sqrt K, \mathbf{x}), \eta^K(\mathbf x )=\sqrt{\frac{K}{K-\|\mathbf{x}\|^{2}}},
\vspace{-0.1in}
\label{l2k}
\end{equation}
where $\eta^K(\mathbf x )$ is the function defining the Lorentz factor, and $(\cdot, \cdot)$ denotes concatenation.

% \begin{figure*}
% \caption{Overall architecture of HVGNN}
% \centering
% \label{}
% \end{figure*}

%\newpage

%!TEX root = ./main.tex

\section{Problem Definition}

In this paper, we consider a dynamic graph where edges evolve over time.
The emergence/disappearance of nodes will lead to the addition/deletion of a set of incident edges concurrently, which can be easily modeled with our proposed approach in a similar way.
Formally, we give the definition of a dynamic graph as follows:

\newtheorem*{def1}{Definition 1 (Dynamic Graph)} 
\begin{def1}
A dynamic graph is defined as a triple $G = (V, E, T)$, where $V = \{v_1,  v_2, $ $ \cdots, v_n\}$ is the node set and each node $v_i$ is associated with a feature vector $\mathbf x_i \in \mathbb R^f$.
$E =\{ (v_i,  v_j )\}$ is the edge set, and $T=\{t_k\}$ is the timestamp set. Each edge $(v_i,  v_j)$ is associated with a timestamp $t_k \in T$ representing that $v_i$ and $v_j$ interact with each other at time $ t_k$.
\end{def1}
%We consider the number of nodes to be a constant. Note that, a node is treated as an isolated one before joining in the graph.
Without loss of generality, we consider the dynamic graph to be attributed.
The timestamps on edges record every interaction among nodes, and thereby fine-grained graph dynamics is captured.
%We are interested in the inductive stochastic representation learning on dynamic graphs in hyperbolic space.
We are interested in inferring stochastic representations in account of graph dynamics and its uncertainty in hyperbolic space.
We prefer to work with the Lorentz model $\mathbb L^{d, K}$ of hyperbolic space owing to its numerical stability and clean closed form expressions with Lorentzian inner product \cite{law2019lorentzian,nickel2018lorentz,HGCN}.
In the Lorentz model, we utilize generalized normal distributions to generate stochastic representations, where the mean and variance are innate to model the uncertainty. 
Formally, we define the problem of representation learning on dynamic graphs as follows: 

\newtheorem*{def2}{Definition 2 (Representation Learning on Dynamic Graphs)} 
\begin{def2}
For a dynamic graph $G = (V, E, T)$, 
the representation learning problem in this paper is to find a map $\Phi: V \to \mathbb L^{d, K}$ 
so that, for each node $v_i$, we can infer stochastic representation $\mathbf z_i(t)$ at any time $t$ in hyperbolic space $\mathbb L^{d, K}$.
The stochastic representation is drawn from a generalized normal distribution in hyperbolic space, 
modeling graph dynamics and its uncertainty.
\end{def2}

%!TEX root = ./main.tex

\section{HVGNN: Hyperbolic Variational GNN}

%We aim to infer stochastic node representations, taking account of graph structure and graph dynamics, in an inductive approach.
%To address this problem, we present a novel Hyperbolic Variational Graph Neural Network (HVGNN).
%In this section, we present a novel HVGNN for dynamic graph representation learning in hyperbolic space.
In a nutshell, HVGNN generates stochastic representations jointly modeling graph dynamics and uncertainty.
% HVGNN outputs the node representation $\mathbf z(t)$ at each time point of a dynamic graph $\mathcal G$,
% where $\mathbf z(t)$ is the random variable of a normal distribution $\mathcal N^\mathbb L(\mathbf \mu(t), \mathbf \sigma(t))$ living in hyperbolic space.
%The overall architecture of HVGNN is illustrated in Fig. 1.
Specifically, to model the dynamics, we propose a novel Temporal Graph Neural Network (TGNN).
To model the uncertainty, we introduce a hyperbolic Variational Graph AutoEncoder (VGAE), 
%whose encoder is defined by the proposed TGNN.
where we utilize TGNN as encoder and give task-oriented decoder for specific task, \eg, link prediction and node classification.
% to generate a random variable $\mathbf z_i(t) \sim \mathcal N^\mathbb L(\mathbf \mu_i(t), \mathbf \sigma_i(t))$ for each node $v_i$, 
% to generate stochastic node representations parameterized by the proposed temporal graph neural network (GNN),
% The aforementioned graph representation learning map $\Phi$ can be defined by the encoder,
% so that we can inductively learn node representations via a single forward pass.
Next, we elaborate on the building blocks of HVGNN, \ie, temporal GNN and hyperbolic VGAE, respectively.

\subsection{Temporal GNN}
We propose the novel TGNN to model graph dynamics in two types of representation spaces: hyperbolic space $\mathbb L^{d, K}$ and Euclidean space $\mathbb R^d$.
Similar to GAT \cite{velickovic2018graph},
the basic idea of TGNN follows the graph attention network.
However, the attention itself cannot handle graph dynamics.
To bridge this gap, the key is to design the following theoretically grounded time encoding approach, distinguishing nodes in time domain.

%Now, we introduce the details of the hyperbolic version of TGNN as follows:

% Similar to GAT and GraphSAGE, 
% The core idea of hyperbolic TGNN is to integrate the attention mechanism and 
% local aggregater. 
% perform time-aware information aggregation.
% Similar to the positional encoding in attention mechanism (...), 
% we propose the hyperbolic time encoding to distinguish the nodes of a temporal graph in the time domain.
% We generalize Euclidean operations to Lorentz model, and propose a masked graph attention in hyperbolic space. 

%dot product to the Lorentzian product $\langle \cdot, \cdot \rangle_\mathcal L$ in hyperbolic space.
% $$\operatorname{Attn}(\mathbf{Q}, \mathbf{K}, \mathbf{V})=\operatorname{softmax}\left(\frac{\mathbf{Q} \mathbf{K}^{\top}}{\sqrt{d}}\right) \mathbf{V}$$
% positional encoding  $\to$ time encoding\\
% attention weights: Euclidean dot product $\to$ Lorentzian product

\subsubsection{Hyperbolic Time Encoding}
We propose a novel time encoding approach to equip graph attention with the ability of modeling dynamics.
% encoding the timestamp, compatible with the attention mechanism.
Formally, we aim to learn a map $\phi_{\mathbb L}: T \to \mathbb L^{d, K}$ from a point in time domain $T$ to a vector in hyperbolic space $\mathbb L^{d, K}$, encoding the temporal information.
The time domain $T$ is $[0, t_{max}]$ and $t_{max}$ denotes the maximum time point in the observed data.

Usually, it is the relative timespan rather than the absolute value of time that reveals critical temporal information.
Thus, in learning the map $\phi_{\mathbb L}: T \to \mathbb L^{d, K}$,
we are interested in the patterns between the relative timespan $|t_i-t_j|$ and the Riemannian metric of hyperbolic space, \ie, Lorentzian inner product. 
%$\langle \phi_{\mathbb L}(t_i) , \phi_{\mathbb L}(t_j) \rangle_\mathcal L$.
We thereby define a Lorentzian kernel $\mathcal K_{\mathbb L}: T \times T \to \mathbb R$ with $\mathcal K_{\mathbb L}(t_i, t_j) = \langle \phi_{\mathbb L}(t_i) , \phi_{\mathbb L}(t_j)  \rangle _{\mathcal L}$.
It is ideal that the kernel can be expressed as a function of relative timespan, namely, \emph{translation invariance}.
Formally, $\mathcal K_{\mathbb L}(t_i, t_j) = \psi_{\mathbb L}(t_i-t_j)$ for some function $\psi: [-t_{max}, t_{max}] \to \mathbb R$.

Instead of investigating $\mathcal K_{\mathbb L}(t_i, t_j) $ explicitly, we propose a two-step approach for hyperbolic time encoding as follows:

\noindent\textbf{Step $1$}: \emph{Construct a translation invariant Euclidean time encoding map.}
We study its Euclidean counterpart  $\mathcal K_{\mathbb R}(t_i, t_j) = \langle \phi_{\mathbb R}(t_i), \phi_{\mathbb R}(t_j) \rangle$ where $\phi_{\mathbb R}: T \to  \mathbb R^d$.
Note that, $\mathcal K_{\mathbb R}$ is positive semidefinite as it is defined by Euclidean inner product.
Accordingly, we have $\mathcal K_{\mathbb R}$ translation invariant, $\mathcal K_{\mathbb R}(t_i, t_j) =\psi_{\mathbb R}(t_i-t_j)$. 
Thus, $\mathcal K_{\mathbb R}$ satisfies the assumption of the \emph{Bochner’s theorem} \cite{Spivak1979}:
% \vspace{-0.18in}
% \begin{abstractbox}
% \vspace{-0.03in}
% A translation-invariant kernel on $\mathbb R^d$ is positive definite iff  there exists a nonnegative measure $p(\omega)$ on $\mathbb R$  such that is the Fourier transform of the measure.
% \vspace{-0.07in}
% \end{abstractbox}
% \vspace{-0.03in}
\emph{A translation-invariant kernel $\mathcal K_{\mathbb R}(t_i, t_j) =\psi_{\mathbb R}(t_i-t_j)$ is positive definite iff  there exists a nonnegative measure $p(\omega)$ on $\mathbb R$  such that $\psi_\mathbb R(\cdot)$ is the Fourier transform of the measure.}

According to the Bochner’s theorem, we have
\begin{equation}
\resizebox{0.905\hsize}{!}{$
\mathcal{K}_\mathbb R \left(t_i, t_j\right)=\int_{\mathbb{R}} e^{\mathbf{i} \omega\left(t_{i}-t_{j}\right)} p(\omega) d \omega=\mathbb{E}_{\omega}\left[\xi_{\omega}\left(t_i\right) \xi_{\omega}\left(t_j\right)^{*}\right],
$}
\label{expection}
\end{equation}
where $\xi_{\omega}(t)=e^{\mathbf{i} \omega t}$, $\mathbf{i}$ is the imaginary unit, and $^*$ denotes the conjugate complex. 
As both kernel $\mathcal K_{\mathbb R}$ and its associated map $\phi_{\mathbb R}$ are real, 
we extract the real part of the expectation in Eq. (\ref{expection}),
% \begin{equation}
% \resizebox{0.905\hsize}{!}{$
% \mathcal{K}_\mathbb R\left(t_i, t_j\right)=\mathbb{E}_{\omega}\left[\cos \left(\omega t_i\right) \cos \left(\omega t_j\right)+\sin \left(\omega t_i\right) \sin \left(\omega t_j\right)\right]
% $}
% \label{exp}
% \end{equation}
which can be approximated by Monte Carlo integral \cite{rahimi2007random}, \ie, 
\begin{equation}
\resizebox{1\hsize}{!}{$
\mathcal{K}_\mathbb R\left(t_i, t_j\right) \approx \frac{1}{d} \sum\limits_{i=1}^d [\cos \left(\omega_{i} t_i\right) \cos \left(\omega_{i} t_j\right)+\sin \left(\omega_{i} t_i\right) \sin \left(\omega_{i} t_j\right)].
$}
\end{equation} 
According to the formulation above, we define  $\phi_{\mathbb R}$ below,
\begin{equation}
\resizebox{1\hsize}{!}{$
\phi_{\mathbb R}(t)= \sqrt{\frac{1}{d}}\left(\cos \left(\omega_{1} t + \theta_1\right), \cos \left(\omega_{2} t + \theta_2\right), \ldots, \cos \left(\omega_{d} t + \theta_d\right) \right),
$}
\end{equation}
where $(\cdot, \cdot)$ denotes the concatenation, and we will learn the $\omega$s and $\theta$s with different subscripts as model parameters.

\noindent\textbf{Step $2$}: \emph{Project Euclidean time encoding to the hyperbolic space.}
We refer to $(\sqrt K, 0, \cdots, 0)$ as the \emph{origin} of Lorentz model, denoted as $\mathcal O$. 
For $\mathbf x \in \mathbb R^d$, we have $(0, \mathbf x)$ live in the tangent space of the origin $\mathcal T_\mathcal O \mathbb L^{d, K}$, since $\langle \mathcal O, (0, \mathbf x) \rangle_\mathcal L=0$. 
Then, $(0, \mathbf x)$ can be projected onto the $\mathbb L^{d, K}$  via the exponential map. 
Finally, we define the projection as 
\begin{equation}
\pi^K_{\mathbb R \to \mathbb L}(\mathbf x) =\exp^K_\mathcal O\left((0, \mathbf x )\right) , 
\label{r2l}
\end{equation}
and obtain $\phi_\mathbb L(t)= \pi^K_{\mathbb R \to \mathbb L}(\phi_\mathbb R(t))$, \ie, 
\begin{equation}
\resizebox{1\hsize}{!}{$
\phi_{\mathbb L}(t) = \left(\sqrt{K} \cosh \left(   \sqrt{\frac{1}{K}} \left\|\phi_{\mathbb R}(t)\right\|  \right), \frac{\sqrt{K}}{\| \phi_{\mathbb R}(t) \|} \sinh \left( \sqrt{\frac{1}{K}} \left\| \phi_{\mathbb R}(t) \right\| \right) \phi_{\mathbb R}(t) \right).
$}
\end{equation}

Now, we prove the translation invariance of  Lorentzian kernel $\mathcal K_{\mathbb L}(t_i, t_j) $ with the hyperbolic time encoding map $\phi_\mathbb L$.
\newtheorem*{thm}{Theorem 1 (Translation Invariance)}
\begin{thm}
The Lorentzian kernel $\mathcal K_{\mathbb L}(t_i, t_j) = \langle  \phi_{\mathbb L}(t_i) , \phi_{\mathbb L}(t_j) \rangle _{\mathcal L}$ with the proposed  $\phi_\mathbb L(\cdot)$ is translation invariant, \ie,
$\mathcal K_{\mathbb L}(t_i, t_j) = \psi_{\mathbb L}(t_i-t_j)$.
\end{thm}
\begin{proof}
We prove the translation invariance of the Lorentzian kernel $\mathcal K_{\mathbb L}(t_i, t_j)$ by proving the existence of the function $\psi_{ \mathbb L}$.
Expanding the Lorentzian product with the definition given in Eq. (\ref{lproduct}), we have the following equation hold:
\begin{equation}
\resizebox{0.7\hsize}{!}{$
\langle  \phi_{\mathbb L}(t_i) , \phi_{\mathbb L}(t_j) \rangle _{\mathcal L}= A \langle  \phi_{\mathbb R}(t_i) , \phi_{\mathbb R}(t_j) \rangle + B,
$}
\end{equation}
where
\begin{equation}
\resizebox{0.93\hsize}{!}{$
\begin{aligned}
A &= -K \sinh \left(\frac{\phi_{ \mathbb R}(t_i)}{\sqrt K} \right) \sinh \left( \frac{\phi_{ \mathbb R}(t_j)}{\sqrt K}\right) \frac{1}{\| \phi_{ \mathbb R}(t_i) \|  \|  \phi_{ \mathbb R}(t_i) \|},\\
B&= -K \cosh \left( \frac{\phi_{ \mathbb R}(t_i)}{\sqrt K}\right) \cosh \left(\frac{\phi_{ \mathbb R}(t_j)}{\sqrt K} \right),
\end{aligned}
$}
\end{equation} 
According to the Bochner’s theorem, we have $\mathcal K_{\mathbb R}(t_i, t_j) = \langle \phi_{\mathbb R}(t_i) , \phi_{\mathbb R}(t_j) \rangle = \psi_{\mathbb R}(t_i-t_j)$. 
Thus, for given $t_i$ and $t_j$,
\vspace{-0.03in}
$$\mathcal K_{\mathbb L}(t_i, t_j) = \langle  \phi_{\mathbb L}(t_i) , \phi_{\mathbb L}(t_j) \rangle _{\mathcal L}=\psi_{\mathbb L}(t_i-t_j),$$
where $\phi^{\mathbb L}=f \circ \psi^{\mathbb R} $ and $f(x)=Ax+B$. 
\end{proof}

\subsubsection{Hyperbolic Operators} 
%We introduce the hyperbolic operators required by hyperbolic TGNN.
In TGNN, we generalize aggregation, addition and linear transformation from Euclidean space $\mathbb R^d$ to hyperbolic space $\mathbb L^{d, K}$.
% The combination in $\mathbb L^{d, K}$ is nontrivial as  addition is not defined in Lorentz model.
% Alternatively, we define the hyperbolic combination $\oplus^K $, combing a hyperbolic $\mathbf x \in \mathbb L^{d, K}$ with an Euclidean $\mathbf y \in \mathbb R$, as follows:
% \begin{equation}
% \mathbf x \oplus^K \mathbf y=\exp^K_{\mathbf x}\left( P_{\mathbf 0 \to \mathbf x}((0, \mathbf y))\right)
% \label{hadd}
% \end{equation}
% Eq. (\ref{hadd}) describes an intuitive procedure: we construct the $(0, \mathbf y)$ in the tangent space of the origin $\mathbf 0$, parallel transport $(0, \mathbf y)$ from $\mathbf 0$ to $\mathbf x$, and project it back onto the $\mathbb L^{d, K}$.
Aggregation is to calculate a weighted midpoint in Euclidean space \cite{HAN}.
%However, it is nontrivial in hyperbolic space. 
%From the geometry perspective, aggregation can be regarded as a midpoint in Euclidean space (..).
%We employ Einstein midpoint to generalize Euclidean midpoint in hyperbolic space.
The Euclidean midpoint is generalized to Einstein midpoint in hyperbolic space.
%Aggregating a set of $\left\{\mathbf{x}_{i}\right\}$ with corresponding attentions $\left\{\alpha_{i}\right\}$, 
Specifically, we have a vector set  $\{\mathbf x_i| \mathbf x_i \in \mathbb L^{d,K}, i \in \Omega\}$, where $\Omega$ is the index set and each $\mathbf x_i$ is associated with a weight $\alpha_i$.
Their Einstein midpoint $\textsc{Agg}^K\left(\left\{\alpha_i, \mathbf{x}_i \right\}_{i \in \Omega}\right)$ is calculated as
\vspace{-0.05in}
\begin{equation}
\textsc{Agg}^K\left(\left\{\alpha_i, \mathbf{x}_i \right\}_{i \in \Omega}\right)=\sum\nolimits_{i}\left[\frac{\alpha_i \eta^K\left(\mathbf{x}_{i}\right)}{\sum_{\ell} \alpha_{\ell} \eta^K\left(\mathbf{x}_{ \ell}\right)}\right] \mathbf{x}_{i},
\vspace{-0.03in}
\end{equation}
where $\eta^K(\cdot)$ is defined in Eq. (\ref{l2k}), and $\mathbf x$'s are the coordinates in Klein model. 
Fortunately, different models of hyperbolic space are essentially the same. 
We can transform coordinates between Klein and Lorentz model via Eqs. (\ref{k2l}) and (\ref{l2k}).
%Additionally, we regard the weighted addition $\oplus^{K}$ as a special case of aggregation, and we have  $\mathbf{x}_{1} \oplus^{K} \mathbf{x}_{2}=\textsc{Agg}^K(\{\alpha_i, \mathbf{x}_i \}_{i \in \ \{1,2\} })$.
Additionally, we regard the weighted addition $\oplus^{K}$ as its special, and have  $\mathbf{x}_{1} \oplus^{K} \mathbf{x}_{2}=\textsc{Agg}^K(\{\alpha_i, \mathbf{x}_i \}_{i \in \ \{1,2\} })$.
%where there are two elements in the set $\left\{\mathbf{x}_{i}\right\}$.
%Then, the weighted addition in Lorentz model  is defined as 
% \vspace{-0.05in}
% \begin{equation}
% \mathbf{x}_{1} \oplus^{K} \mathbf{x}_{2}= \frac{\beta \eta^K\left(\mathbf{x}_{1}\right)\mathbf{x}_{1} +(1- \beta) \eta^K\left(\mathbf{x}_{2}\right)\mathbf{x}_{2} }
% {\beta \eta^K\left(\mathbf{x}_{1}\right) +(1-\beta) \eta^K\left(\mathbf{x}_{2}\right) },
% \vspace{-0.05in}
% \end{equation}
% \vspace{-0.05in}
% \begin{equation}
% \mathbf{x}_{1} \oplus^{K} \mathbf{x}_{2}=\textsc{Agg}^K(\{\alpha_i, \mathbf{x}_i \}_{i \in \ \{1,2\} }).
% \vspace{-0.05in}
% \end{equation}
%where $\beta \in (0, 1)$ is the weight.

%Then, we introduce the linear transformation in hyperbolic space. 
Linear transformation in hyperbolic space is realized by matrix vector multiplication, denoted as $\otimes^K$.
We define $\otimes^K$ via exponential and logarithmic maps as follows:
\vspace{-0.05in}
\begin{equation}
\mathbf W \otimes^{K} \mathbf{x}=\exp _{\mathcal O}^{K}\left(\mathbf W \log _{\mathcal O}^{K}\left(\mathbf{x}\right)\right), 
\vspace{-0.05in}
\label{lt}
\end{equation}
where $\mathbf W$ is the weight matrix. 
The intuition is that we perform linear transformation in the Euclidean tangent space and then map the result back onto hyperbolic space.

\begin{figure}
\centering
    \includegraphics[width=0.93\linewidth]{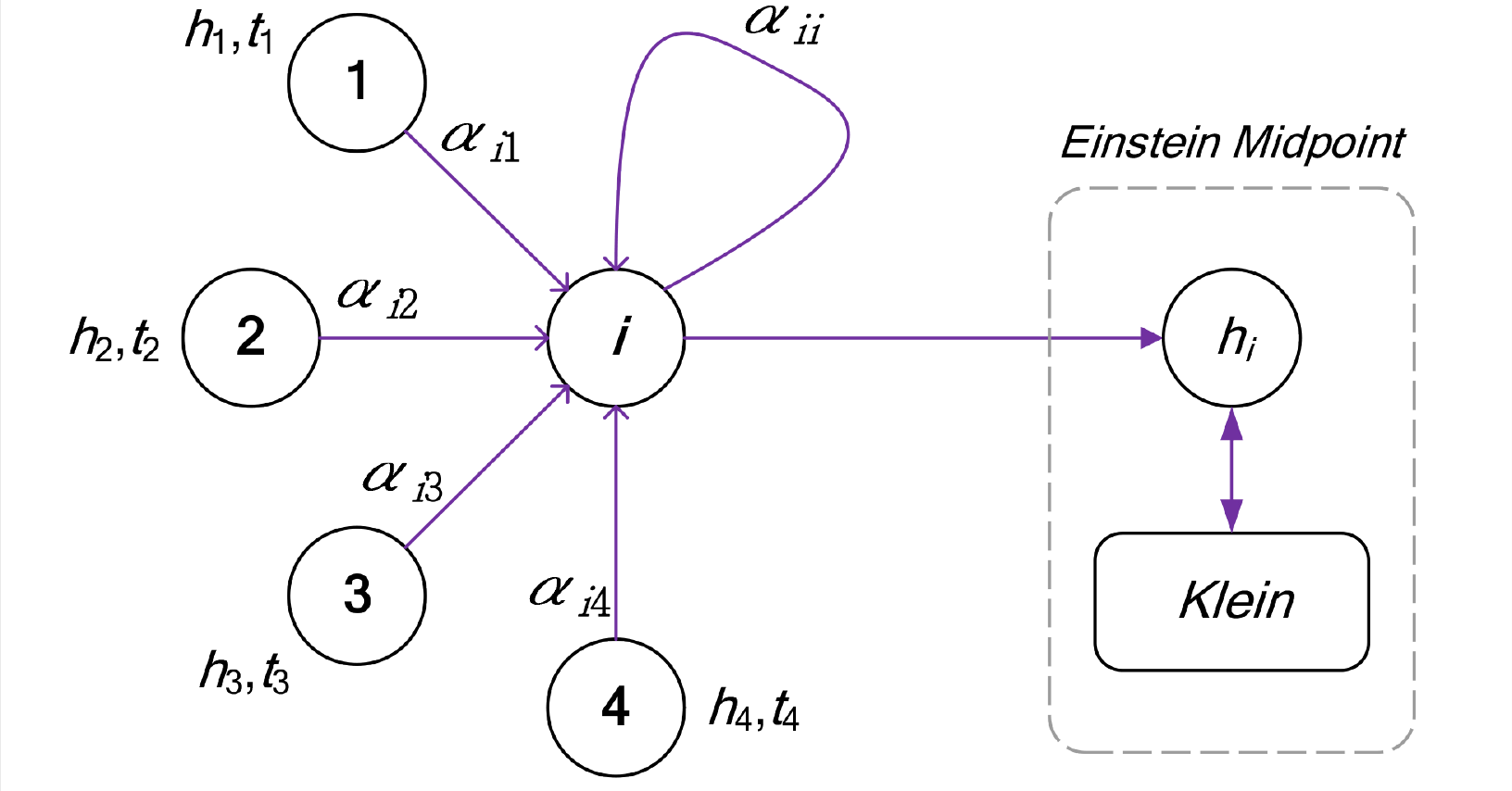}
    \vspace{-0.1in}
        % \caption{\textsc{HypTGA} Layer. It preforms aggregation over time-aware neighborhood. The aggregation  is obtained by calculating Einstein midpoint with the aid of Klein model.}
               \caption{\textsc{HypTGA} Layer. In the time-aware neighborhood, the timestamps $t_1, \cdots, t_4$ are prior to the given time $t$.  Einstein midpoint is calculated with the aid of Klein model.}
    \label{tgnn}
\end{figure}

\subsubsection{Hyperbolic Temporal Graph Attention} 

% To this end, we combine hyperbolic time encoding $\mathbf t^\mathbb L \in \mathbb L^{d,K}$ with feature vector $\mathbf v \in \mathbb R^d$ as raw node vector $\mathbf v = \mathbf t^\mathbb L  \oplus^K \mathbf x$.

% We propose a novel hyperbolic TGNN, denoted as $\operatorname{TGNN}^\mathbb L$, to perform time-aware information aggregation.
% Similar to GAT, we design $\operatorname{TGNN}^\mathbb L$ as a local aggregater, however, $\operatorname{TGNN}^\mathbb L$ take accounts of  graph structure as well as graph evolvement in hyperbolic space $\mathbb L^{d, K}$.

We denote hyperbolic TGNN in hyperbolic space $\mathbb L^{d, K}$ as $\operatorname{TGNN}_\mathbb L$. 
We build $\operatorname{TGNN}_\mathbb L$ by stacking its sole building block layer, \ie, the \emph{hyperbolic temporal graph attention} (\textsc{HypTGA}) layer. 
The \textsc{HypTGA} layer aims to renew node representations at time point $t$, modeling graph dynamics.

As opposed to static graph attention (\eg, GAT) receiving all the neighbors' feature, we conduct aggregation in account of the interaction time between the neighbors.
%imestamps associated with the edge.
% We denote the output representation for node $v_i$ at time $t$ from the $l^{th}$ layer as $\mathbf h_i^{(l)}(t)$.
% We introduce the definition of temporal neighborhood before giving its input.
Specifically, for a target node $v_i$ at time $t$, we define its time-aware neighborhood $N_{i, t}=\{v_1, \cdots, v_N\}$ such that the timestamp $t_j$ of  the interaction between $v_i$ and $v_j \in N_{i, t}$ is prior to $t$. 
\textsc{HypTGA} layer takes node representations $\mathbf h(t)$ and timestamps of  node union $\{v_i \cup N_{i, t}\}$ as the input. 

%The \textsc{HypTGA} layer considers temporal pattern between the target and its neighbors to capture graph dynamics. 
%aggregates the representations together with temporal features, modeling graph structure and graph dynamics. 
%
In addition to graph attention, the \textsc{HypTGA} layer further integrates the time encoding $\phi_\mathbb L (\cdot)$ so that we can distinguish the neighbors in time domain to model graph dynamics.
We focus on relative temporal pattern across the neighbors, which remains the same between a shift. 
Thanks to \emph{translation invariance} of $\phi_\mathbb L (\cdot)$, we can encode $\hat t_j =| t-t_j|$ for each neighbor $v_j$ since $|t_i-t_j|=|(t-t_i)-(t-t_j)|$. 
%Instead of the absolute value, we encode the timespan via $\phi_\mathbb L (\cdot)$ thanks to .
%We incorporate the time relationship and node representation $\mathbf h_i^{(l-1)}(t)$ as follows:
%Specifically, we incorporate $\phi_\mathbb L (\hat t_j)$ with representation $\mathbf h_i$ as follows:
% \vspace{-0.05in}
% \begin{equation}
% \tilde{\mathbf h}_i^{(l)}(t) = \phi_{\mathbb L}^{(l)} (\hat t_i) \oplus^K \mathbf W^{(l)}  \otimes^K \mathbf h_i^{(l-1)}(t),
% \vspace{-0.05in}
% \end{equation}
Then, we obtain the time-aware representation $\tilde{\mathbf h}_j(t)$ as follows:
\vspace{-0.033in}
\begin{equation}
\tilde{\mathbf h}_j(t) = \phi_{\mathbb L} (\hat t_j) \oplus^K \mathbf W \otimes^K \mathbf h_j(t),
\label{combine}
\end{equation}
where $\oplus^K$ and $\otimes^K$ are defined before.
$\otimes^K$ always has a higher priority than $\oplus^K$, similar to their Euclidean counterparts.
%Then, we aggregate $\tilde{\mathbf h}_i$ over the temporal neighborhood,
As shown in Fig. \ref{tgnn}, target node representation $\mathbf h_i(t)$  is updated by aggregating time-aware representations in $N_{i,t}$,
\vspace{-0.033in}
\begin{equation}
\mathbf h_i(t) = \textsc{Agg}^K(\{\alpha_{ij}, \tilde{\mathbf h}_j(t) \}_{j\in \Omega }), \ \Omega =i \cup N_{i, t},
\end{equation}
% \vspace{-0.05in}
% \begin{equation}
% \mathbf h_i = \textsc{Agg}^K(\{\alpha_{ij}, \tilde{\mathbf h}_j \}_{j\in N_{i, t} }) \oplus^K\mathbf h_i,
% \vspace{-0.03in}
% \end{equation}
where we add a self-loop for the target node with $\hat t_j=0$. 
The attention weight $\alpha_{ij}$ is calculated by the attention function $\textsc{Attn}( \cdot, \cdot)$.
Naturally, we define $\textsc{Attn}( \cdot, \cdot)$ by Lorentzian inner product with a nonlinear activation as follows,
\vspace{-0.033in}
\begin{equation}
\textsc{Attn}(\tilde{\mathbf h}_i(t), \tilde{\mathbf h}_j(t))=h(\gamma \langle \tilde{\mathbf h}_i(t), \tilde{\mathbf h}_j(t) \rangle_{\mathcal L}+ c),
\end{equation}
where $\gamma$ and $c$ are weight and bias, respectively.
The bias is placed as Lorentzian inner product is restricted in $\mathbb L^{d, K}$, \ie, $\langle \mathbf x, \mathbf y \rangle_\mathcal L<-K$.
% for $\mathbf x, \mathbf y \in \mathbb L^{d, K}$. 
We define $h(\cdot)$  as $sigmoid(\cdot)$.
% to normalize a scalar.
Recalling Eq. (\ref{lt}), $\operatorname{TGNN}_\mathbb L$ parameters live in the Euclidean tangent space.
Such design will facilitate the model learning.

In $\operatorname{TGNN}_\mathbb L$, for Euclidean input features, we use $\pi^K_{\mathbb R \to \mathbb L}(\cdot)$ in Eq. (\ref{r2l}) to project them onto hyperbolic space $\mathbb L^{d,K}$.
% We project input features to hyperbolic space via exponential map, which is similar to the projection of Euclidean time encoding.
% Formally, we obtain hyperbolic feature $\mathbf h_i^{(0)}(t)=\exp_\mathcal O((0, \mathbf x))$.
Meanwhile, we give its Euclidean counterpart, $\operatorname{TGNN}_\mathbb R$. 
In particular, we utilize Euclidean time encoding $\phi_\mathbb R(\cdot)$, and replace hyperbolic operators of $\operatorname{TGNN}_\mathbb L$ with Euclidean ones.

\subsection{Hyperbolic VGAE}
To model the uncertainty, we introduce a Hyperbolic Variational Graph AutoEncoder (HVGAE) built upon TGNN.
At the time point $t$, HVGAE infers stochastic representations $\mathbf z(t)$ of generalized normal distributions with a variational approach in hyperbolic space $\mathbb L^{d,K}$.
We bridge the gaps of variational approach in hyperbolic space, which are 
%and elaborate on the model architecture  in following subsections.
% stochastic representations take into account graph structure as well as graph evolvement.
% We first introduce a canonical wrapped normal distribution in Lorentz model, and then introduce the architecture of the proposed hyperbolic variational autoencoder.
1) generalizing the (usual) normal distribution, and 2) defining its variational family in hyperbolic space.

 \subsubsection{Wrapped Normal Distribution}
A canonical approach for generalization is to map a usual normal distribution onto hyperbolic space.
Such a probability measure is referred to as wrapped normal distribution \cite{mathieu2019continuous}.
We derive the Probability Density Function (PDF) in $\mathbb L^{d, K}$:
\vspace{-0.033in}
\begin{equation}
\resizebox{0.888\hsize}{!}{$
\begin{aligned}
& \mathcal{N}_{\mathbb L}^{K}(\mathbf{z} \mid \boldsymbol{\mu},  \operatorname{diag}(\boldsymbol \sigma^2)) \\
=&
 F\mathcal{N}([P^K_{\boldsymbol \mu \to \mathcal O }(\mathbf u)]_{-} \mid \mathbf{0}, \operatorname{diag}(\boldsymbol \sigma^2)) 
\sinh (\sqrt{\frac{1}{K}}  \| \mathbf{u} \|_{\mathcal L})^{1-d}
,
\end{aligned}
$}
\label{wnormal}
\vspace{-0.1in}
\end{equation} 
where $F= \left(\frac{ \| \mathbf{u} \|_{\mathcal L}}{\sqrt K}\right)^{d-1}$, 
$\mathbf u = \log^K _{\boldsymbol{\mu}}(\mathbf {z})$
and $[\mathbf x]_{-}=\mathbf x_{2:d+1}$ for $\mathbf x \in \mathbb R^{d+1}$. 
%the derivation details are specified in Supplementary Material.
%Eq. (\ref{wnormal}) recovers the PDF of (usual) normal distribution in the limit when the curvature is set to zero.
The wrapped normal distribution owns two parameters, \ie,
a mean $\boldsymbol \mu \in \mathbb L^{d, K}$ in Lorentz model, 
and a variance $\boldsymbol \sigma \in \mathbb R^{d}$ of a normal distribution $\mathcal{N}(\mathbf{0},\operatorname{diag}(\boldsymbol \sigma^2))$ in Euclidean space.
%most prominent
The advantage of wrapped normal distribution is that the PDF is \emph{differentiable w.r.t. the parameters}, 
so that we can introduce a reparameterisable sampling algorithm, Algorithm $1$, to enable the gradient-based learning.
Specifically, for a wrapped normal sample,
% can be constructed via reparametrisable sampling procedure:
a) we sample a vector from usual normal distribution in Euclidean space (Line $1$), 
b) move the vector to the mean $\boldsymbol{\mu}$ (Line $2$ and $3$), and 
c) map it onto hyperbolic space $\mathbb L^{d,K}$ (Line $4$).
%We give the formal description in Algo. $1$.

\begin{algorithm}
           \caption{Reparametrisable Sampling}
           \LinesNumbered
           \KwIn{parameter $\boldsymbol \mu \in \mathbb L^{d,k}$, $\boldsymbol \sigma \in \mathbb R^d$}
           \KwOut{a sample $\mathbf z \sim \mathcal{N}_{\mathbb L}^{K}(\mathbf{z} \mid \boldsymbol{\mu}, \operatorname{diag}(\boldsymbol \sigma^2))$}
           Sample $\mathbf {\tilde{v}} \sim \mathcal{N}(\mathbf{0},\operatorname{diag}(\boldsymbol \sigma^2)) \in \mathbb R^d$\;
           Construct $\mathbf {v}= (0, \mathbf {\tilde{v}} ) \in \mathcal T_{\mathcal O}\mathbb L^{d, K}$\;  
           %Note: The origin $\mathcal O$ is defined as $(\sqrt K, 0, \cdots, 0)$. \;
           Transport $\mathbf {v}$ to $\mathbf u=P_{\mathcal O \to \boldsymbol \mu}(\mathbf{ v}) \in \mathcal T_{\boldsymbol \mu}\mathbb L^{d, K}$ via Eq. (\ref{pt})\;
           Map $\mathbf u$ to $\mathbf z=\exp_{\boldsymbol \mu}(\mathbf u) \in \mathbb L^{d,K}$ via Eq. (\ref{exp}).
\end{algorithm}

\subsubsection{Hyperbolic Variational Family} In HVGAE, we need to define a prior distribution and the family of corresponding variational posterior in hyperbolic space.
In particular, the prior is the standard distribution $\mathcal{N}_{\mathbb L}^{K}(\cdot|\mathbf 0, \mathbf I)$, 
The variational family is $\{ \mathcal{N}_{\mathbb L}^{K}(\cdot|\boldsymbol{\mu}, \operatorname{diag}(\boldsymbol \sigma^2)) \ | \boldsymbol \mu \in \mathbb L^{d, K}, \boldsymbol \sigma \in \mathbb R^{d}\}$, where we propose to define the parameters as
$\boldsymbol \mu=\operatorname{TGNN}_\mathbb L(G)$ and
$\log \boldsymbol \sigma=  \operatorname{TGNN}_\mathbb R(G)$.
%In this way, the generated stochastic representations are able to encode the dynamics and uncertainty.
\emph{The novelty is two-fold}: 
1) We define the distribution parameters by the time-aware TGNN, so that the uncertainty is jointly modeled with the dynamics.
2) We introduce a reparametrisable sampling algorithm to enable the gradient-based learning.
%by the proposed $\operatorname{TGNN}$s

\subsection{Overall Architecture}
\begin{figure}
\centering
    \includegraphics[width=0.97\linewidth]{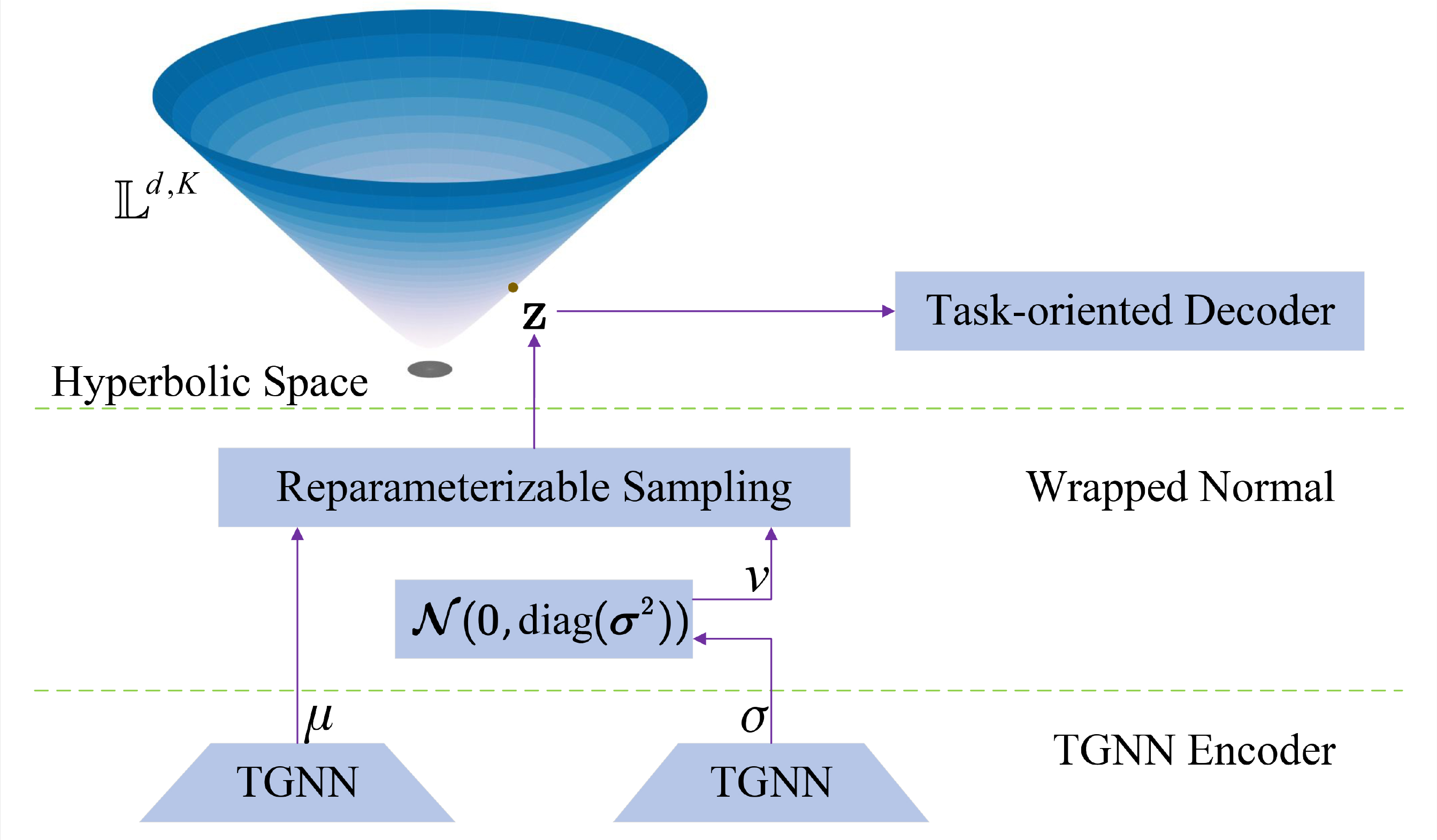}
        \vspace{-0.1in}
        \caption{The overall architecture of HVGNN.
        % In the hyperbolic representation space $\mathbb L^{d,K}$,
        % HVGNN incorporates the hyperbolic VGAE to infers stochastic representations of generalized normal distributions in $\mathbb L^{d,K}$, 
        % whose parameters are defined by TGNN encoders. 
        % The hyperbolic $\operatorname{TGNN}_\mathbb L$ and Euclidean $\operatorname{TGNN}_\mathbb R$ output the mean $\boldsymbol \mu$ and the variance $\boldsymbol \sigma$, respectively. 
        %We utilize the TGNN encoder and Fermi-Dirac decoder, and introduce a reparameterisable sampling to enable gradient-based learning.
        }
    \label{architecture}
\end{figure}

%In order to generate the node representation $\mathbf z \sim \mathcal{N}_{\mathbb L}^{K}(\mathbf{z}\mid\boldsymbol{\mu}, \operatorname{diag}(\boldsymbol \sigma^2))$,
Thanks to reparameterisable sampling, HVGNN can be learned in an end-to-end approach with a specific learning task. 
HVGNN is built with an encoder-decoder framework, illustrated in Fig. \ref{architecture}, and the aforementioned representation learning map $\Phi$ can be defined by the encoder.
In particular, we utilize TGNN encoder and a task-oriented decoder.
In this paper, we provide the decoder for link prediction and node classification.
%Next, we introduce the encoder, decoder and learning objective.

% We build the hyperbolic VGAE with $\operatorname{TGNN}$ encoder and link prediction decoder, 
% parameterizing the variational posterior and defining the likelihood, respectively.

\subsubsection{Encoder}
For node $v_i$ at time $t$, we denote the stochastic representation as $\mathbf z_i(t) \in \mathbb L^{d,K}$, summarized in $\mathbf Z_t$, and denote the class label as $y_i=k \in [1, C]$, summarized in $\mathcal Y$.
With the TGNN encoder, we give the posterior as follows:
%Given $\mathcal G$, the posterior $q(\mathbf{Z}(t)|\mathcal{G})$ is given as follows:
\begin{equation}
q(\mathbf{Z}_t \mid G)=\prod\nolimits_{i=1}^{n} q\left(\mathbf{z}_{i}(t) \mid G\right),
\end{equation} 
where  $n$ is the number of nodes in the graph.   
$q\left(\mathbf{z}_{i}(t)|G\right)$ is defined by the hyperbolic variational family above.
%$\mathcal{N}_{\mathbb L}^{d,K}(\mathbf{z}_i(t) \mid \boldsymbol{\mu}, \operatorname{diag}(\boldsymbol \sigma^2))$.

\subsubsection{Decoder}
%We provide decoder for link prediction and node classification.
For the task of \emph{link prediction}, we utilize the Fermi-Dirac decoder. Formally, we have 
\begin{equation}
p(G|\mathbf{Z}_t)=\prod\nolimits_{(v_i, v_j) \in E}  p\left((v_i, v_j) \in E|\mathbf{z}_{i}(t), \mathbf{z}_{j}(t)\right),
\end{equation}
whose likelihood is defined by the Fermi-Dirac function \cite{HGCN}. 
For \emph{node classification}, we utilize the hyperbolic multinomial logistic decoder, formally,
\vspace{-0.05in}
\begin{equation}
p(\mathcal{Y}|\mathbf{Z}_t)=\prod\nolimits_{i=1}^{n}  p\left( y_i=k \mid \mathbf{z}_{i}(t) \right),
\end{equation}
whose likelihood is defined by the hyperbolic multinomial logistic function \cite{HNN}.
% $$p\left((v_i, v_j) \in \mathcal E|\mathbf{z}_{i}(t), \mathbf{z}_{j}(t)\right)=\left( e^{(\langle\mathbf{z}_{i}(t), \mathbf{z}_{j}(t)\rangle_{\mathcal L}-s)/\kappa} +1 \right)^{-1}.$$
% It is a generalized sigmoid function with parameters $(s, \kappa)$.

\subsubsection{Learning objective}
We formulate the learning objective following the vanilla Variational AutoEncoder \cite{2014Auto}.
The decoder is trained together with the encoder by evidence lower bound (ELBO) defined as follows:
\begin{equation}
\mathcal{J}=\mathbb{E}_{q(\mathbf{Z}_t \mid G)}[\log p(\cdot \mid \mathbf{Z}_t)]-\mathrm{KL}[q(\mathbf{Z}_t \mid G) \| p(\mathbf{Z}_t)],
\end{equation}
where $p(\cdot \mid \mathbf{Z}_t)$ is the likelihood of corresponding decoder. 
$\mathrm{KL}[ \cdot \| \cdot ]$ is the Kullback-Leibler (KL) divergence between posterior $q(\mathbf{Z}_t \mid G)$ and prior $p(\mathbf{Z}_t)=\prod_{i} \mathcal{N}_{\mathbb L}^{K}(\mathbf z_i(t) | \mathbf 0, \mathbf I)$. 
The reparametrisable sampling enables the evaluation of the gradient of the ELBO \wrt  of $\operatorname{TGNN}$ parameters living in Euclidean spaces.
Consequently, we can make use of usual optimizer to learn the model.
With the learned model, stochastic representations can be inferred inductively modeling graph dynamics and uncertainty in hyperbolic space.

%allows us to evaluate the ELBO exactly and to take the gradient of the objective function
% we can compute the density of the probability distribution.

% This parametric formulation of q is called
% reparametrization trick, and it enables the evaluation of the
% gradient of the objective function with respect to the network
% parameters.

%!TEX root = ./main.tex

\begin{table*}[h]
  %\scriptsize
    \centering
 % \resizebox{1\textwidth}{!}{
        \begin{tabular}{ p{2.5cm}<{\centering}| p{2.05cm}<{\centering} p{2.05cm}<{\centering}| p{2.05cm}<{\centering} p{2.05cm}<{\centering}| p{2.05cm}<{\centering} p{2.05cm}<{\centering} }
        \hline
                    & \multicolumn{2}{c|}{\textbf{Reddit}} &\multicolumn{2}{c|}{\textbf{Wikipedia}} &\multicolumn{2}{c}{\textbf{DBLP}}\\
        \textbf{ Model }&Accuracy &AP &Accuracy &AP &Accuracy &AP  \\
        \hline
        GAT                &$90.54\pm0.18$     & $95.05\pm0.21$    &$ 87.05\pm0.33$     & $ 94.52\pm0.20$    & $88.24 \pm0.54$    &$ 94.71\pm0.28$  \\
        GraphSAGE    &$91.21\pm0.22$     & $95.12\pm 0.16$   &$85.21\pm0.28$      &$93.16\pm0.25$      & $87.36\pm0.12$     &$94.08\pm0.35$  \\
        TGAT              &$92.52\pm0.25$     & $96.11\pm0.20$    &$87.65\pm0.24$      &$95.02\pm0.08$      & $89.11\pm0.46$     &$95.46\pm0.17$  \\
        HGCN             &$92.03\pm0.41$     &$95.86\pm0.08$     &$87.12\pm0.33$      &$94.67\pm0.38$      &$88.83\pm0.45$      &$95.18\pm0.14$  \\
        \hline
        TGNN$_\mathbb R$       &$92.18\pm0.06$    &$96.03\pm0.27$    &$88.11\pm0.47$    &$95.36\pm0.32$    &$89.25\pm0.32$   &$95.67\pm0.29$  \\
        TGNN$_\mathbb L$        &$93.36\pm0.31$    &$97.25\pm0.07$    &$89.74\pm0.08$    &$96.62\pm0.22$    &$91.12\pm0.51$   &$97.12\pm0.36$  \\
                \hline
        EVGNN                           &$93.20\pm0.27$     &$97.37\pm0.35$    &$90.05\pm0.62$    &$96.85\pm0.39$    &$90.89\pm0.06$   &$97.18\pm0.14$  \\
       \textbf{HVGNN}             &$\mathbf{94.72\pm0.32}$     &$\mathbf{98.79\pm0.48}$    &$\mathbf{91.67\pm0.40}$     &$\mathbf{98.02\pm0.15}$    &$\mathbf{92.17\pm0.21}$   &$\mathbf{98.25\pm0.09}$  \\
        \hline
    \end{tabular}
   %}
       \caption{The performance of \emph{transductive} link prediction in terms of Accuracy and AP ($\%$).}
   \label{tab:tlp}
\end{table*}

\begin{table*}[h]
    \centering
  %  \resizebox{1\textwidth}{!}{
            \begin{tabular}{ p{2.5cm}<{\centering}| p{2.05cm}<{\centering} p{2.05cm}<{\centering}| p{2.05cm}<{\centering} p{2.05cm}<{\centering}| p{2.05cm}<{\centering} p{2.05cm}<{\centering} }
         \hline
                    & \multicolumn{2}{c|}{\textbf{Reddit}} &\multicolumn{2}{c|}{\textbf{Wikipedia}} &\multicolumn{2}{c}{\textbf{DBLP}}\\
          \textbf{ Model }&Accuracy &AP &Accuracy &AP &Accuracy &AP  \\
          \hline
          GAT               &$88.12\pm0.22$      &$93.37\pm0.28$   &$ 82.14\pm0.31$    &$ 91.12\pm0.39$    &$84.26 \pm0.12$   & $ 91.88\pm0.61$  \\
          GraphSAGE   &$87.63\pm0.15$      &$94.64\pm0.24$   & $82.26\pm0.42$    &$90.86\pm0.33$     &$83.75\pm0.26$    &$92.35\pm0.39$  \\
          TGAT             &$89.25\pm0.21$     &$95.12\pm0.33$    &$85.05\pm0.18$     &$93.51\pm0.27$     &$86.38\pm0.51$    &$94.47\pm0.36$  \\
          HGCN            &$89.07\pm0.29$     &$95.18\pm0.26$    &$84.93\pm0.49$     &$93.18\pm0.35$     &$86.41\pm0.29$    &$94.02\pm0.26$  \\
          \hline
          TGNN$_\mathbb R$       &$89.08\pm0.40$    &$94.87\pm0.57$    &$85.63\pm0.26$    &$93.95\pm0.23$    &$86.24\pm0.02$   &$94.13\pm0.41$  \\
          TGNN$_\mathbb L$       &$90.34\pm0.20$    &$96.11\pm0.39$    &$87.05\pm0.38$    &$95.14\pm0.19$    &$88.60\pm0.45$   &$95.35\pm0.17$  \\
          \hline
          EVGNN                           &$90.27\pm0.03$    &$95.93\pm0.57$    &$87.52\pm0.48$    &$95.08\pm0.42$    &$88.48\pm0.10$   &$95.22\pm0.35$  \\
          \textbf{HVGNN}            &$\mathbf{91.89\pm0.06}$    &$\mathbf{97.67\pm0.20}$    &$\mathbf{89.13\pm0.11}$   &$\mathbf{97.15\pm0.45}$    &$\mathbf{90.33\pm0.07}$   &$\mathbf{97.36\pm0.13}$  \\
          \hline
   %\scriptsize
    \end{tabular}
                \caption{The performance of \emph{inductive} link prediction in terms of Accuracy and AP ($\%$).}
    \label{tab:ilp}
   % }
      \vspace{-0.05in}
\end{table*}

\begin{table*}[h]
    \centering
  %  \resizebox{1\textwidth}{!}{
        \begin{tabular}{ p{2.5cm}<{\centering}| p{2.05cm}<{\centering} p{2.05cm}<{\centering} p{2.05cm}<{\centering}| p{2.05cm}<{\centering} p{2.05cm}<{\centering} p{2.05cm}<{\centering} }
         \hline
                    & \multicolumn{3}{c|}{\textbf{Transductive Setting}} &\multicolumn{3}{c}{\textbf{Inductive Setting}}\\
       \textbf{ Model }& \textbf{Reddit} &\textbf{Wikipedia} &\textbf{DBLP} &\textbf{Reddit} &\textbf{Wikipedia} &\textbf{DBLP}  \\
        \hline
        GAT                 &$64.12\pm0.48$      & $81.54\pm0.80 $     &$78.48\pm1.29$      &$ 62.55\pm0.61$    &$79.18 \pm0.12$   &$77.03\pm0.55$  \\
        GraphSAGE    &$60.86\pm0.59$       & $82.05\pm0.72$      &$77.67\pm0.05$      &$59.63\pm0.47$     &$79.67\pm0.31$   &$76.12\pm0.19$  \\
        TGAT              &$64.88\pm 0.67$      &$83.28\pm0.56$       &$78.73\pm0.12$      &$63.24\pm0.33$     &$80.05\pm0.80$   &$77.86\pm0.36$  \\
        HGCN              &$64.67\pm0.42$      &$82.96\pm0.13$       &$79.42\pm0.45$       &$63.98\pm0.16$    &$80.12\pm0.92$   &$77.51\pm0.23$  \\
        \hline
        TGNN$_\mathbb R$     &$64.53\pm0.14$    &$83.19\pm0.49$    &$78.79\pm0.72$     &$63.17\pm0.39$     &$79.89\pm0.61$    &$77.95\pm0.47$  \\
        TGNN$_\mathbb L$      &$65.92\pm0.54$    &$84.55\pm0.96$    &$80.05\pm0.33$    &$64.03\pm0.25$     &$81.27\pm0.35$    &$79.16\pm0.29$  \\
        \hline
        EVGNN                         &$66.35\pm0.69$    &$84.71\pm0.15$    &$81.13\pm0.24$    &$65.18\pm0.61$    &$82.02\pm0.54$   &$80.24\pm0.56$  \\
        \textbf{HVGNN}          &$\mathbf{68.13\pm0.51}$    &$\mathbf{86.22\pm0.42}$    &$\mathbf{82.67\pm0.14}$    &$\mathbf{67.26\pm0.75}$    &$\mathbf{83.96\pm0.24}$   &$\mathbf{81.72\pm0.58}$  \\
        \hline
  %  \scriptsize
    \end{tabular}
   % }
     % \vspace{-0.07in}
     \caption{The performance of \emph{transductive} and \emph{inductive} node classification in terms of AUC ($\%$).}
   \label{tab:nc}
 %  \vspace{-0.07in}
  \end{table*}

\section{Experiments}
We evaluate HVGNN by link prediction and node classification on several datasets. 
We repeat each experiment 10 times and report the mean with the standard deviations.

\subsection{Experimental Setups}
\subsubsection{Datasets} 
We choose three real-world datasets, \ie, \textbf{Reddit} \cite{Xu2020inductive}, \textbf{Wikipedia} \cite{kumar2019predicting} and \textbf{DBLP} \cite{zhou2018sparc}.
% \begin{itemize}
%   \item \textbf{Reddit} \cite{Xu2020inductive}: We collect active users and their posts under subreddits, yielding a dynamic graph with $11,000$ nodes and $703,055$ timestamped edges.
%   \item \textbf{Wikipedia} \cite{kumar2019predicting}: We collect top edited pages and active users, yielding a dynamic graph of $9,300$ nodes and $160,572$ timestamped edges.
%   \item \textbf{DBLP} \cite{zhou2018sparc}: It is a citation network from several CV conferences, yielding a dynamic graph of $1,909$ nodes and $8,237$ edges with publication time as timestamps.
% \end{itemize}
In Reddit, we collect active users and their posts under subreddits, yielding a dynamic graph with $12,000$ nodes and $763,055$ timestamped edges.
In Wikipedia, we collect top edited pages and active users, yielding a graph of $9,300$ nodes and $160,572$ timestamped edges.
DBLP is a citation network from several CV conferences, yielding a  graph of $1,909$ nodes and $8,237$ edges with publication time as timestamps.
\subsubsection{Comparison Method}
% We compare the proposed model, HVGNN, against both the Euclidean models and hyperbolic model, which can be evaluated in both transductive and inductive settings.
%Most existing models learn the representations transductively, however, inductive learning is important especially for dynamic graphs.
We compare the proposed model, HVGNN, against the models that can be evaluated in both transductive and inductive settings.
%both the Euclidean models and hyperbolic model, which
In particular, Euclidean models include two baselines for static graphs, \ie, \textbf{GAT} \cite{velickovic2018graph} and \textbf{GraphSAGE} \cite{hamilton2017inductive}, and a recent one for dynamic graphs, \ie, \textbf{TGAT} \cite{Xu2020inductive}.
Hyperbolic model includes \textbf{HGCN} \cite{HGCN}, a recent model for static graphs.
These methods generate deterministic representations and thereby cannot capture uncertainty. 
% \begin{itemize}
% 	% \item EvolveGCN: It captures graph dynamics by using an RNN to evolve the GCN parameters.
% 	% \item VGRNN: It is a hierarchical variational model with additional latent stochastic variables for modeling dynamic graphs.
% 	\item \textbf{TGAT} \cite{Xu2020inductive}: It combines graph attention and a time encoding method in Euclidean space. 
% 	\item \textbf{HGCN} \cite{HGCN}: It introduces hyperbolic operators, and generalizes a graph convolutional network for static graphs in hyperbolic space.
% \end{itemize}
% The methods above generate deterministic representations. TGAT has a similar idea with the proposed TGNN. However, it lacks the ability to generalize to hyperbolic space.
%Except TGAT, these methods model static graphs neglecting the dynamics.

\subsubsection{Ablation Study} 
We include the proposed dynamic graph models, \ie, $\operatorname{TGNN}_\mathbb R$ and $\operatorname{TGNN}_\mathbb L$, without modeling the uncertainty as baselines. 
Additionally, we design the Euclidean variant of HVGNN, namely EVGNN, which models dynamic graphs with the uncertainty in Euclidean space. 
Specifically, we utilize two $\operatorname{TGNN}_\mathbb R$ to parameterize usual normal distributions. EVGNN is then optimized with reparameterisable trick of the standard VAE.
On the one hand, we can study the importance of modeling uncertainty by comparing HVGNN (EVGNN) against $\operatorname{TGNN}_\mathbb L$ ($\operatorname{TGNN}_\mathbb R$). On the other hand, we can study the effect of the representation space by comparing hyperbolic models against their Euclidean counterparts.
In the experiment, we stack the corresponding attention layer twice in the models above.
%HVGNN, EVGNN, $\operatorname{TGNN}_\mathbb L$ and $\operatorname{TGNN}_\mathbb R$.
Refer Supplementary Material for further experimental details.
%To further evaluate the superiority of hyperbolic space, 

%To evaluate the effectiveness of hyperbolic time encoding approach, we equip the hyperbolic model (HGCN) for static graph with this approach, and refer to the equipped models as HGCN+T. Specifically, we add the hyperbolic encoding and the hyperbolic node feature via Eq. (\ref{combine}) before aggregation.

\subsubsection{Transductive and Inductive Settings}
%We evaluate the performances under transductive and inductive settings.
The transductive setting examines output representations of the nodes that have been observed in training.
The inductive setting examines output representations of \emph{unseen} nodes while training.
Node representations are initialized as its raw feature.
We do chronological train-validation-test split with $80\%-5\%-15\%$ according to the timestamps.

\subsection{Link Prediction}
The task of link prediction is to predict the probability of two nodes being connected.
%by an edge. 
%We respect the representation spaces of graph models. 
For hyperbolic models, we utilize the Fermi-Dirac decoder with Lorentz inner product to compute the probability based on the learned representations. 
For Euclidean models, we replace the Lorentz inner product with normal inner product.
The graph models are trained by minimizing the cross-entropy loss using negative sampling.
%HVGNN owns an additional KL divergence loss to regulate stochastic representations drawn from the wrapped normal distributions.
We randomly sample an equal amount of negative node pairs to the positive links, and employ the and classification \emph{Accuracy} and \emph{Average Precision} (\emph{AP}) as evaluation metrics.
The performances under transductive and inductive settings are reported in Tables \ref{tab:tlp} and \ref{tab:ilp}, respectively.
% \begin{itemize}
% \item The proposed model, HVGNN, consistently outperforms its competitors in both transductive and inductive settings. 
% \item HGCN+T beats HGCN in general. This suggests that in hyperbolic space we can equip a static model with hyperbolic time encoding to enable dynamic graph modeling. 
% \item HVGNN achieves better performance than TGNN. This shows the importance to model the uncertainty.
% \item Hyperbolic models (HVGNN and TGNN$_\mathbb L$) consistently outperforms the Euclidean counterparts (EVGNN and TGNN$_\mathbb R$). The reasons lies in that hyperbolic space is well-suited for modeling real-world graphs, which is also supported in the literature. 
% \end{itemize}
The proposed model, HVGNN, consistently outperforms its competitors. 
% 2) HGCN+T beats HGCN in general. This suggests that in hyperbolic space we can equip a static model with hyperbolic time encoding to enable dynamic graph modeling. 
For example, on Wikipedia dataset, HVGNN obtains $3\%$ and $3.64\%$ performance gains against its best competitor in transductive and inductive settings, respectively.
The reason is that HVGNN models the inherent characteristics of graphs, \ie, the dynamics and uncertainty, in the promising hyperbolic space.
Additionally, we provide further insights though the ablation study:
1) Hyperbolic models (HVGNN and $\operatorname{TGNN}_\mathbb L$) outperform the Euclidean counterparts (EVGNN and $\operatorname{TGNN}_\mathbb R$). 
This suggests hyperbolic space is a more promising representation space well-suited for modeling real-world graphs. 
%2) HVGNN compares favorably to HGCN showing the importance of modeling dynamics.
2) HVGNN and EVGNN achieve better performance than $\operatorname{TGNN}_\mathbb L$ and $\operatorname{TGNN}_\mathbb R$, respectively. 
This shows that the uncertainty cannot be ignored for modeling graphs.

\subsection{Node Classification}
The task of node classification is to predict the label of the node based on node representations.
% generated by the graph models.
%Respecting the representation spaces, 
We utilize usual multinomial logistic loss for Euclidean models,    
while the hyperbolic multinomial logistic loss for hyperbolic models.
We train the comparison models together with the link prediction loss above.
%, encouraging node representations to preserve the graph structure.
Owing to the label imbalance on the datasets, we employ the \emph{area under the ROC curve} (\emph{AUC}) as the evaluation metric.
We summarize the performances under transductive and inductive settings in Table ~\ref{tab:nc}.
As reported in Table ~\ref{tab:nc}, HVGNN achieves the best performance consistently. 
For example, on Wikipedia dataset, HVGNN outperforms its best competitor by $2.94\%$ and $3.84\%$ in transductive and inductive settings, respectively.
Additionally, 
%the effectiveness of hyperbolic time encoding,
the superiority of hyperbolic space and the importance of modeling dynamics and uncertainty are supported in the experimental results.

\begin{figure} 
\centering 
\subfigure[Link Prediction]{
\includegraphics[width=0.48\linewidth]{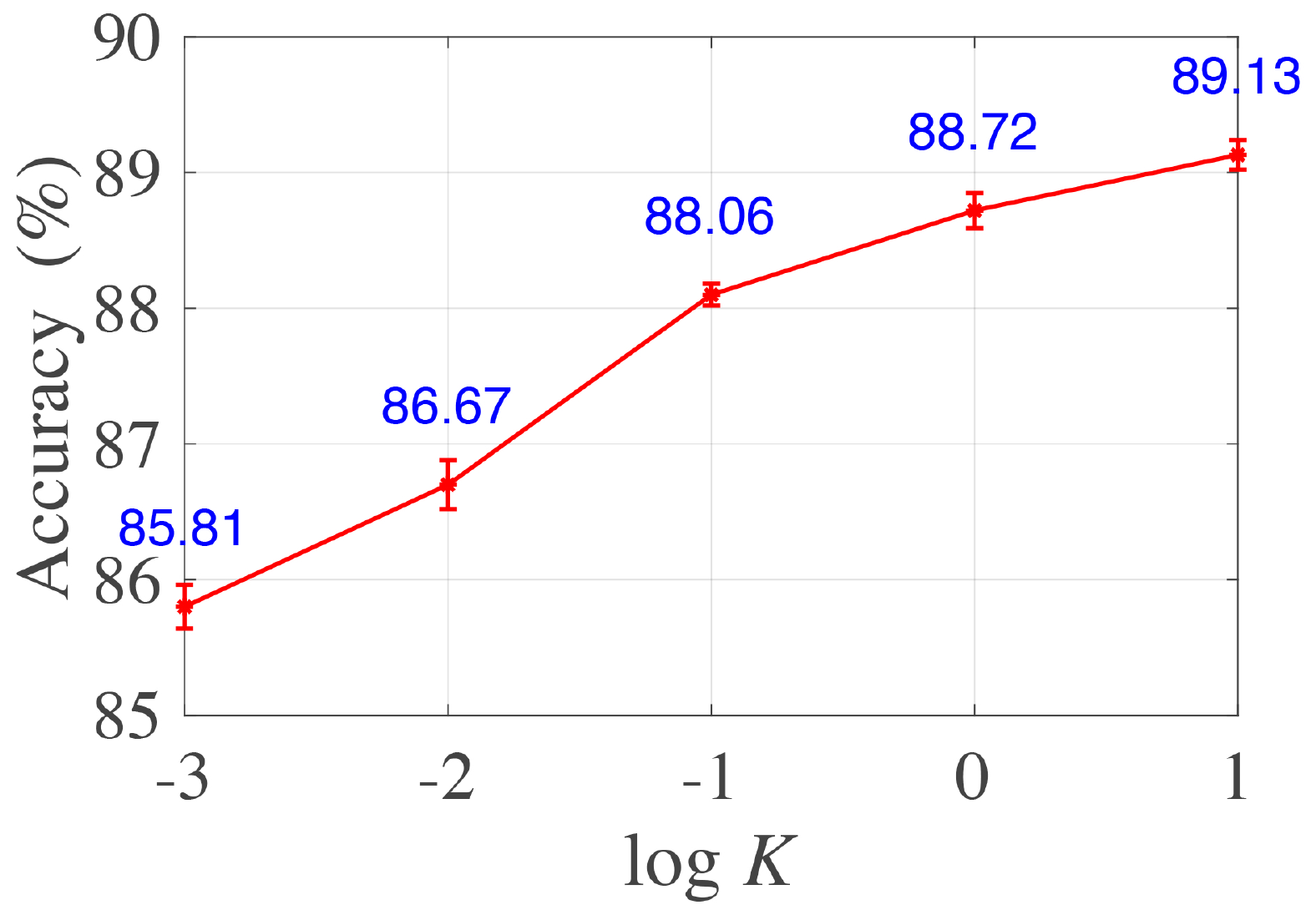}}
%\hspace{0.01\linewidth}
\subfigure[Node Classification]{
\includegraphics[width=0.48\linewidth]{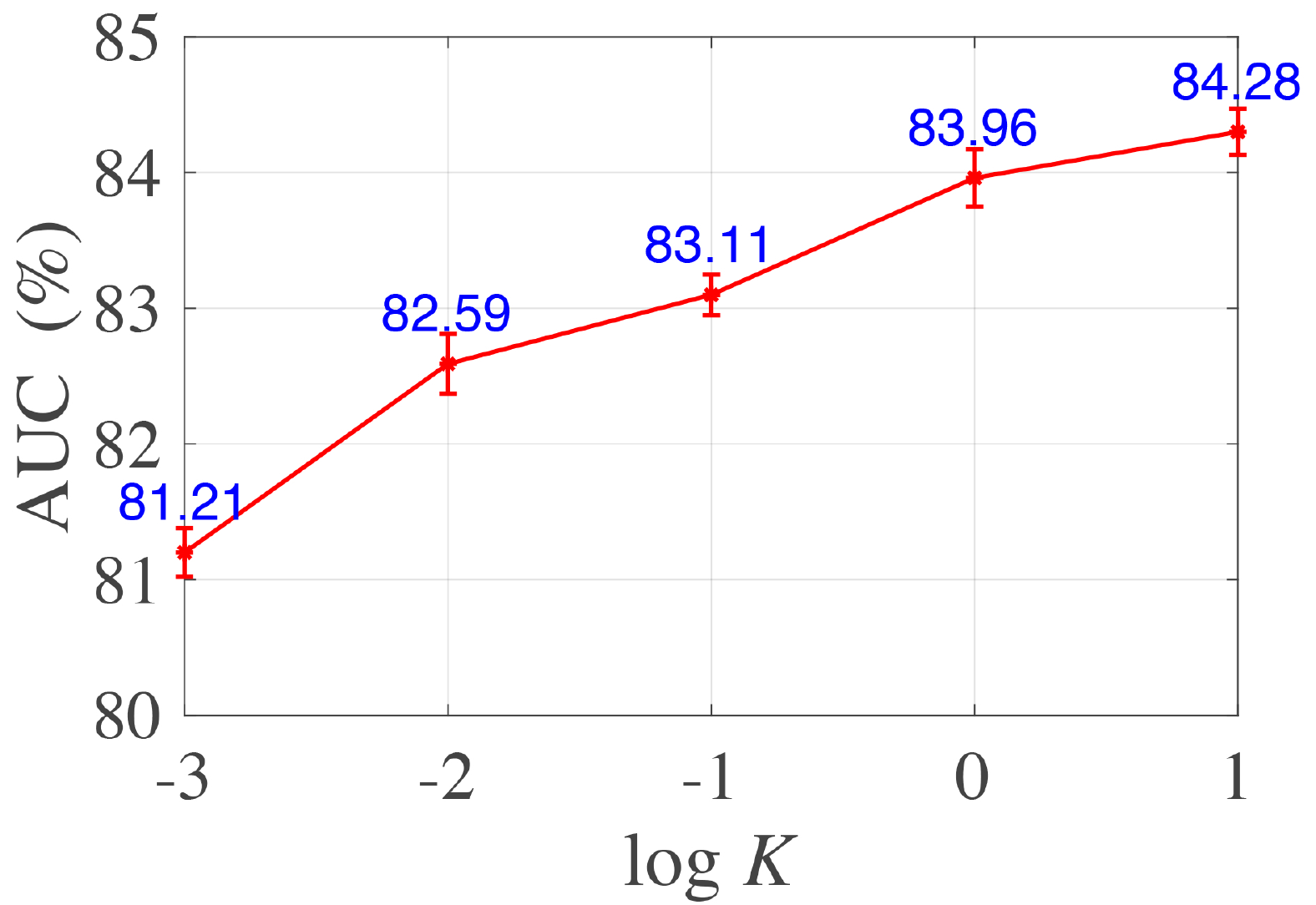}}
\vspace{-0.1in}
\caption{The effects of curvature $-1/K$ for inductive learning on Wikipedia dataset. }
%( The shadow shows deviations.)}
\vspace{-0.12in}
\label{figure:curvature}
\end{figure}

\subsection{The Curvature of Hyperbolic Space}
In HVGNN, we consider the curvature of hyperbolic space, controlled by $K$, as a hyperparameter.
In fact, the hyperbolic spaces with different curvatures are essentially the same \cite{Spivak1979}.
However, owing to the limited machine precision and normalization, hyperbolic spaces with different curvatures lead to different performances \cite{HGCN}.
For instance, as shown in Figure~\ref{figure:curvature}, we obtain performance gain by adjusting the curvature.
% This motivates us to make the design decision in HVGNN that, instead of a fixed curvature, we give the generalized formulation of a trainable curvature to improve its learning capacity. 
Motivated by this observation, instead of a fixed curvature, in HVGNN, we give the generalized formulation of a trainable curvature to improve its learning capacity.

%!TEX root = ./main.tex

\section{Related Work}
%We briefly summarize the related work in following areas:
% and thereby actives strong classifiers and clustering methods \cite{2021mCRF,2019XiaClassify}, e.g., a state-of-the-art k-means with no bounds \cite{2020A}, requiring vector inputs.
\subsection{Graph Representation Learning}
Graph representation learning generates vector representations for graphs, 
and thereby activates advances in machine learning with vector inputs, e.g., a recent strong classifier \cite{2019XiaClassify} whose inputs are coarse granular features.
Thus, it is attracting increasing attentions, and finds itself from network alignment \cite{sun2019dna,sun2018ijcai} to text classification \cite{2019Hierarchical}.
% Prior works model graphs via random walk \cite{DeepWalk}, or define explicit optimization objectives \cite{tang2015line,}.
Among graph modeling methods, GNNs \cite{wang2016structural,ma2019Disentangled,kipf2016semi} play an important role.
In general, graph modeling is widely studied in static settings while models for dynamic graphs are still scant. 
Recently, several solutions \cite{zhou2018dynamic,Xhonneux2020CGNN} for dynamic graphs are proposed.
For example, EvolveGCN \cite{pareja2019evolvegcn} models dynamics with a sequence of snapshots.
% TGAT \cite{Xu2020inductive} generates node representations for dynamic graph in an inductive fashion.
JODIE \cite{kumar2019predicting} models the node trajectories.
VGRNN \cite{2019Variational} further models the uncertainty.
To our knowledge, all prior models for dynamic graphs consider node representations in Euclidean space, 

\subsection{Hyperbolic Representation Learning}
Most of representation learning methods assume the representation space to be Euclidean.
%Recently, researchers are concerned with the representation space itself.
Actually, hyperbolic space provides an exciting alternative.
It is well-suited to model hierarchical data \cite{krioukov2010hyperbolic,papadopoulos2012popularity}.
An increasing number of studies report hyperbolic model compares favorably to its Euclidean counterpart in a wide spectrum of applications, such as word embedding \cite{nickel2017poincare,tifrea2018poincare}, question answering \cite{tay2018hyperbolic}, clustering \cite{monath2019gradient}, network alignment \cite{sun2020perfect} and reasoning in knowledge graph \cite{balazevic2019multi}.
Most of existing methods generate deterministic vectors living hyperbolic space, while some recent works \cite{mathieu2019continuous,nagano2019wrapped} study the generalized normal distributions for stochastic representations.
Recently, hyperbolic graph models \cite{HNN,HGNN} have been proposed where the graph is considered to be static.
Distinguishing from these studies, we propose the first hyperbolic model for dynamic graphs.

%!TEX root = ./main.tex

% In this paper, we propose to learn the representations for dynamic graphs in hyperbolic space, for the first time.
% To this end, we present a novel HVGNN, 
% which inductively infer stochastic representations to model graph structure and its uncertainty.
% In HVGNN, 
% %we work with Lorentz model of hyperbolic space for its numerical stability.
% % we first propose the TGNN, where we empower the attention mechanism to handle graph dynamics via a novel time encoding approach.
% we first introduce the TGNN based on a novel time encoding approach.
% Then, we propose the HVGAE built upon TGNN to infer stochastic representations of wrapped normal distributions.
% Furthermore, we introduce a reparameterisable sampling to enable the gradient-learning of HVGNN.

\section{Conclusion}

We have presented a novel HVGNN for dynamic graph representation learning in hyperbolic space.
HVGNN captures graph dynamics and uncertainty in the stochastic representations of wrapped normal distributions in hyperbolic space.
HVGNN further incorporates a reparameterisable sampling algorithm to enable its gradient-based learning.
Experimental results show the superiority of HVGNN for link prediction and node classification on several real-world datasets.

\newpage

\section{Acknowledgments}
% This work was supported in part by: National Natural Science Foundation under Grant U1936103 and 61921003, the National Key Research and Development Program of China under Grant 2018YFB1003804, Fundamental Research Funds for the Central Universities 2019XD11, National Science Foundation under grants III-1763325, III-1909323, IIS-1763365 and SaTC-1930941.
This work was supported by the National Key Research and Development Program of China under Grant 2018YFB1003804, National Natural Science Foundation of China under Grant U1936103, 61921003 and 62002007, and Fundamental Research Funds for the Central Universities 2019XD11.
Philip S. Yu was supported by National Science Foundation under grants III-1763325, III-1909323 and SaTC-1930941. Jiawei Zhang was supported by National Science Foundation under grants IIS-1763365.

%\section{Ethics Statement}
\section{Broader Impact}
This paper introduced a novel research problem: dynamic graph representation learning in hyperbolic space. 
This paper broadens the current graph representation learning research in multiple dimensions, \eg, hyperbolic representation space, dynamic graphs, and stochastic representation learning, to new stages and greatly enriches the current graph neural network studies. 
This paper possibly motivates further studies on representation learning in the Riemannian manifolds in a more general and elegant way.
Furthermore, the model (HVGNN) proposed in this paper has transformative impacts in various real-world applications and a wide spectrum of interdisciplinary studies on graph data, such as online social computing, recommender systems, bio-medical studies, and neural science. 
Both results and source code will be released to the public, and others who otherwise have limited access to the models can use our open-source materials in their researches or applications.
We would encourage researchers to explore further applications of our approach, and also welcome the discussion on any theoretical and empirical details, and all kinds of improvements and enhancements from any research field.

\bibliographystyle{aaai}
\bibliography{aaai21}
% discovery sciences.
% In this study, we obey AAAI Publication Ethics and Malpractice Statement.
% For the first time, we propose to learn dynamic graph representation in hyperbolic space.
% The proposed approach models the inherent dynamics and uncertainty of graphs in the promising hyperbolic space,
% and extensive experiments show the superiority of the proposed approach on several real-world datasets.
% The proposed approach shows the potentials to support a wide spectrum of downstream applications, e.g., user-item interaction networks for recommendation, biochemical networks for medical benefits and financial transaction networks for social goods.
% This study sheds lights on introducing differential geometry methodologies to graph mining.

\end{document}

\maketitle
In the technical appendix, we provide further details for derivation and proof as well as experiments.

\section{Transition Invariance}
We first review the notion of \emph{transition invariance} and then prove the theorem in Section of Temporal GNN in details. 
%of the theorem in Section 4.
\newtheorem*{def1}{Definition 1} 
\begin{def1}
A kernel $\mathcal K(t_i, t_j): \mathbb R \times \mathbb R \to \mathbb R$ is said to be transition invariance iff there exists a function that $\mathcal K(t_i, t_j)=\psi(t_i- t_j).$
\end{def1}
The notion of transition invariance is intuitive. For a transition $C$, we have the kernel $\mathcal K$ invariance, i.e., $\mathcal K(t_i+C, t_j+C)=\psi((t_i+C)-(t_j+C))=\psi(t_i- t_j)=\mathcal K(t_i, t_j)$.
\newtheorem*{thm1}{Theorem 1}
\begin{thm1}
In our paper, the Lorentzian kernel $\mathcal K_{\mathbb L}(t_i, t_j) = \langle  \phi_{\mathbb L}(t_i) , \phi_{\mathbb L}(t_j) \rangle _{\mathcal L}$ with the proposed  $\phi_\mathbb L(\cdot)$ is translation invariant, i.e.,
$\mathcal K_{\mathbb L}(t_i, t_j) = \psi_{\mathbb L}(t_i-t_j)$.
\end{thm1}
\begin{proof}
We prove the translation invariance of the Lorentzian kernel $\mathcal K_{\mathbb L}(t_i, t_j)$ by proving the existence of the function $\psi_{ \mathbb L}$.
Expanding the Lorentzian product with the definition, we have the following equation hold:
%\vspace{-0.05in}
\begin{equation}
\langle  \phi_{\mathbb L}(t_i) , \phi_{\mathbb L}(t_j) \rangle _{\mathcal L}= A \langle  \phi_{\mathbb R}(t_i) , \phi_{\mathbb R}(t_j) \rangle + B,
\end{equation}
where
\begin{equation}
\begin{aligned}
A &= -K \sinh \left(\frac{\phi_{ \mathbb R}(t_i)}{\sqrt K} \right) \sinh \left( \frac{\phi_{ \mathbb R}(t_j)}{\sqrt K}\right) \frac{1}{\| \phi_{ \mathbb R}(t_i) \|  \|  \phi_{ \mathbb R}(t_i) \|},\\
B&= -K \cosh \left( \frac{\phi_{ \mathbb R}(t_i)}{\sqrt K}\right) \cosh \left(\frac{\phi_{ \mathbb R}(t_j)}{\sqrt K} \right),
\end{aligned}
\end{equation} 
where $K$ is the constant defining the curvature of hyperbolic space.
According to the Bochner’s theorem, we have $\mathcal K_{\mathbb R}(t_i, t_j) = \langle \phi_{\mathbb R}(t_i) , \phi_{\mathbb R}(t_j) \rangle = \psi_{\mathbb R}(t_i-t_j)$.  
Furthermore, based on the study \cite{Xu2020inductive}, for a reasonable finite dimension $d$ in practice, the kernel $\mathcal K_{\mathbb R}$ can be approximated with arbitrary low error $\epsilon>0$, i.e., 
\begin{equation}
\sup _{t_{1}, t_{2} \in T}\left| \langle \phi_{\mathbb R}\left(t_{i}\right), \phi_{\mathbb R}\left(t_{j}\right)\rangle -\mathcal{K}\left(t_{i}, t_{j}\right)\right|<\epsilon
\end{equation}
Thus, for given $t_i$ and $t_j$,
$\mathcal K_{\mathbb L}(t_i, t_j) = \langle  \phi_{\mathbb L}(t_i) , \phi_{\mathbb L}(t_j) \rangle _{\mathcal L}=\psi_{\mathbb L}(t_i-t_j)$,
where $\phi^{\mathbb L}=f \circ \psi^{\mathbb R} $ and $f(x)=Ax+B$. Additionally, the kernel can be constructed by a finite time encoding generated by $\phi_{\mathbb L}(t)$ with arbitrary low error. 
\end{proof}

\section{Wrapped Normal Distribution}
We elaborate on the density function of wrapped normal distribution in the Lorentz model of hyperbolic space $\mathbb L^{d,K}$, denoted as $\mathcal{N}_{\mathbb L}^{K}(\boldsymbol z \mid \boldsymbol{\mu},  \Sigma)$.

As specified in Algorithm $1$, a wrapped normal distribution is constructed by mapping a normal distribution in Euclidean space onto hyperbolic space $\mathbb L^{d,K}$. 
The map is formulated as $f^K_{\boldsymbol\mu}:=\exp^K _{\boldsymbol{\mu}} \circ P^K_{\mathcal O \rightarrow \boldsymbol{\mu}}$.
Then, the density function of wrapped normal distribution can be derived utilizing the theorem \cite{2010Measure} below:

%According to the measure theory, we have the theorem below:
\newtheorem*{thm2}{Theorem 2}
\begin{thm2}
If  $f(\cdot)$ is an invertible and differentiable map, and $\boldsymbol{z}$ is a random variable endowed with the probability density function $p(\boldsymbol{x})$, the log likelihood of $\boldsymbol{z} =
f(\boldsymbol{x})$  can be expressed as
\begin{equation}
\log p(\boldsymbol{z})=\log p(\boldsymbol{x})-\log \operatorname{det}\left(\frac{\partial f}{\partial \boldsymbol{x}}\right),
\label{xtoz}
\end{equation}
where $\operatorname{det}(\cdot)$ is the determinant.
\end{thm2}

First, we study the property of $f^K_{\boldsymbol \mu}$.
Recall the closed-form expression of parallel transport and exponential map.
For any $\boldsymbol \nu, \boldsymbol \mu \in \mathbb L^{d,K}$, we have $ (P^K_{\boldsymbol \nu \rightarrow \boldsymbol \mu})^{-1}(\boldsymbol{u})=P^K_{\boldsymbol{\mu} \rightarrow \boldsymbol \nu}(\boldsymbol{u})$.
For $\boldsymbol{z}=\exp^K _{\boldsymbol{\mu}}(\boldsymbol{u})$, we have $\boldsymbol{u}=(\exp ^K_{\boldsymbol{\mu}})^{-1}(\boldsymbol{z})= \log^K_{\boldsymbol{\mu}}(\boldsymbol{z})$. 
In fact, logarithmic map serves as the inverse operator of exponential map, and vice versa.
Additionally, it is easy to check that both parallel transport and exponential map are differentiable.
Hence, $f^K_{\boldsymbol\mu}=\exp^K _{\boldsymbol{\mu}} \circ P^K_{\mathcal O \rightarrow \boldsymbol{\mu}}$ is an invertible and differentiable map.
$p(\boldsymbol x)$ in Eq. (\ref{xtoz}) is the density function of usual normal distribution in Euclidean space, 
\begin{equation}
\mathcal N(\boldsymbol x| \boldsymbol u, \mathbf \Sigma)=\frac{1}{\sqrt{ (2\pi)^d \operatorname{det} \mathbf \Sigma }} e^{-\frac{1}{2}( \boldsymbol x- \boldsymbol u)^\top\mathbf \Sigma^{-1}( \boldsymbol x- \boldsymbol u)},
\end{equation}
where we have $\boldsymbol x= (f^K_{\boldsymbol\mu})^{-1}(\boldsymbol z)$ and set $\boldsymbol u = \mathbf 0$ according to Algorithm $1$.

Next, the challenge lies in the determinant of the differential of $f^K_{\boldsymbol \mu}$.
Appealing to the chain-rule and the property of determinant, we can decompose the expression into two components as follows:
\begin{equation}
\operatorname{det} \left(\frac{\partial f^K_{\boldsymbol \mu}(\boldsymbol{v})}{\partial \boldsymbol{v}}\right) 
=\operatorname{det}\left(\frac{\partial \exp^K _{\boldsymbol{\mu}}(\boldsymbol{u})}{\partial \boldsymbol{u}}\right) \cdot \operatorname{det}\left(\frac{\partial P^K_{\mathcal O \rightarrow \boldsymbol{\mu}}(\boldsymbol{v})}{\partial \boldsymbol{v}}\right).
\end{equation}
Parallel transport itself is norm preserving, and this means $\operatorname{det}\left(\frac{\partial }{\partial \boldsymbol{v}} P^K_{\mathcal O \rightarrow \boldsymbol \mu}(\boldsymbol{v})  \right)=1$.
According to the study \cite{nagano2019wrapped}, we obtain the determinant of the differential of the exponential map as follows:
\begin{equation} 
\operatorname{det}\left(\frac{\partial \exp^K _{\boldsymbol{\mu}}(\boldsymbol{u})}{\partial \boldsymbol{u}}\right)=\left(\frac{\sinh ( \sqrt K \|\boldsymbol{\mu}\|_{\mathcal{L}})}{ \sqrt K \|\boldsymbol{\mu}\|_{\mathcal{L}}}\right)^{d-1}.
\end{equation}
Finally, we give the density function $\mathcal{N}_\mathbb {L}^{K}(\boldsymbol z \mid \boldsymbol{\mu},  \mathbf \Sigma) $ below:
\begin{equation}
\mathcal{N}(P^K_{\mathcal O \to \boldsymbol \mu}( \log^K _{\boldsymbol{\mu}}(\boldsymbol z) ) \mid \mathbf{0}, \mathbf \Sigma)\left(\frac{\sqrt{K} \| \boldsymbol{\mu} \|_{\mathcal L}}{\sinh (\sqrt{K} \| \boldsymbol{\mu} \|_{\mathcal L})}\right)^{d-1},
\label{wnormal}
\end{equation} 
where the covariance matrix $\mathbf \Sigma$ is set to $\operatorname{diag}(\boldsymbol \sigma^2)$ for the easy of parameterization.

\section{Experimental Details}
In this section, we give further experimental details to enhance \emph{reproducibility}.
\subsection{Code and Data}
The datasets used in this paper, i.e., \textbf{Reddit} \cite{Xu2020inductive}, \textbf{Wikipedia} \cite{kumar2019predicting} and \textbf{DBLP} \cite{zhou2018sparc}, are publicly available.
We submit the source code and will publish the code after acceptance.  
\subsection{Running Environment}
All experiments were conducted on the a CentOS server with two $11$G Nvidia RTX 2080Ti and $128$G RAM.
\subsection{Implementation Notes}
\begin{itemize}
\item For comparison models, \textbf{GAT} \cite{velickovic2018graph}, \textbf{GraphSAGE} \cite{hamilton2017inductive}, \textbf{TGAT} \cite{Xu2020inductive}, \textbf{HGCN} \cite{HGCN}, we utilize  the code released by authors and the hyperparameters of best performance.
\item For the variants, EVGNN, $\operatorname{TGNN}_\mathbb L$ and $\operatorname{TGNN}_\mathbb R$, they can be implemented based on the code of  the proposed model, HVGNN.
\end{itemize}
\subsection{Parameter Settings}
For the architecture of the proposed neural model, we stack the corresponding attention layer twice in $\operatorname{TGNN}_\mathbb R$, $\operatorname{TGNN}_\mathbb L$, EVGNN and HVGNN. 
We find that we cannot obtain obvious performance gain when stacking more attention layer.
For $\operatorname{TGNN}_\mathbb L$ and HVGNN, the dimensionality of hyperbolic space is $32$ in the reported results. 
For $\operatorname{TGNN}_\mathbb L$ and EVGNN, the dimensionality of Euclidean space is set to be the same as their Euclidean competitors for the fairness.
In this paper, the dimensionality of Euclidean space is $256$.
As specified in the section of (The Curvature of Hyperbolic Space), we evaluate the parameter sensitivity of $K$. In particular, $\log K$ varies in $[-3, -2, -1, 0, 1]$.
% \newpage

\bibliographystyle{aaai}
\bibliography{aaai21}